\documentclass[acmsmall]{acmart}

\AtBeginDocument{%
  \providecommand\BibTeX{{%
    \normalfont B\kern-0.5em{\scshape i\kern-0.25em b}\kern-0.8em\TeX}}}
    
\usepackage[utf8]{inputenc}
\DeclareUnicodeCharacter{0301}{\'{e}}
\usepackage{geometry}
\usepackage{amsthm}
\usepackage{graphicx}
\usepackage{multirow}
\usepackage{pifont}
\usepackage{array}
\usepackage{svg}
\usepackage{float}
\usepackage{subcaption}
\newtheorem{definition}{Definition}
\title{Linkage on Security, Privacy and Fairness in Federated Learning: New Balances and New Perspectives}

\author{Linlin Wang}
\email{linlinwang.cityu@gmail.com}
\affiliation{%
  \institution{City University of Macau}
  \streetaddress{Avenida Padre Tomás Pereira Taipa}
  \city{Macau}
  \country{China}
}

\author{Tianqing Zhu*}
\email{tianqing.zhu@uts.edu.au}
\affiliation{%
  \institution{University of Technology Sydney}
  \streetaddress{PO Box 123 Broadway}
  \city{Sydney}
  \state{NSW}
  \country{Australia}
  \postcode{2007}
}

\author{Wanlei Zhou}
\email{wlzhou@cityu.edu.mo}
\affiliation{%
  \institution{City University of Macau}
  \streetaddress{Avenida Padre Tomás Pereira Taipa}
  \city{Macau}
  \country{China}
}
\author{Philip S. Yu}
\email{psyu@uic.edu}
\affiliation{%
  \institution{University of Illinois at Chicago}
  \streetaddress{851 S. Morgan St., Rm 1138 SEO, Chicago, IL 60607}
  \city{Chicago}
  \country{US}
}
\begin{abstract}

Federated learning is fast becoming a popular paradigm for applications involving mobile devices, banking systems, healthcare, and IoT systems. Hence, over the past five years, researchers have undertaken extensive studies on the privacy leaks, security threats, and fairness associated with these emerging models. For the most part, these three critical concepts have been studied in isolation; however, recent research has revealed that there may be an intricate interplay between them. For instance, some researchers have discovered that pursuing fairness may compromise privacy, or that efforts to enhance security can impact fairness. These emerging insights shed light on the fundamental connections between privacy, security, and fairness within federated learning, and, by delving deeper into these interconnections, we may be able to significantly augment research and development across the field. Consequently, the aim of this survey is to offer comprehensive descriptions of the privacy, security, and fairness issues in federated learning. Moreover, we analyze the complex relationships between these three dimensions of cyber safety and pinpoint the fundamental elements that influence each of them. We contend that there exists a trade-off between privacy and fairness and between security and gradient sharing. On this basis, fairness can function as a bridge between privacy and security to build models that are either more secure or more private. Building upon our observations, we identify the trade-offs between privacy and fairness and between security and fairness within the context of federated learning. The survey then concludes with promising directions for future research in this vanguard field.

\end{abstract}

\begin{document}

\maketitle

\section{Introduction}
Federated learning has garnered considerable attention in recent years due to its potential to effectively address data privacy concerns among multiple parties~\cite{yang2019federated}. Federated learning is a distributed learning approach that comprises local clients and a central server~\cite{mcmahan2017communication}. Within this framework, the local clients each train their own machine learning model using their own local data, while the central server coordinates their activities. Each client then sends the parameters of their model to the central server, who aggregates the updates and returns a global model to each client~\cite{issa2023blockchain}. Crucially, the central server never directly accesses the local data, which drastically reduces the risk of information leaks and privacy violations. Since its inception, federated learning has found numerous applications across a diverse range of fields, including mobile devices~\cite{lim2020federated}, banking systems~\cite{yang2019federated}, healthcare institutions ~\cite{nguyen2022federated}, the Internet of Things (IoT)~\cite{rahman2021challenges}, and other domains ~\cite{hu2016energy}. Furthermore, federated learning has been explored as a potential solution to address the issue of data fragmentation, commonly referred to as the "data island problem" \cite{yang2019federated}.

Although federated learning was originally designed to safeguard machine learning privacy, previous research has revealed a myriad of privacy and security vulnerabilities inherent to federated learning.  Concerns over fairness have also been raised~\cite{mothukuri2021survey}. For instance, several studies~\cite{yin2021comprehensive,mothukuri2021survey,zhang2023fedrecovery,zhou2022adversarial} have explored how adversaries and attacks, such as gradient leaks and inference attacks, can introduce privacy risks to a federated learning framework. Manifest security issues include poisoning attacks and backdoor attacks. In addition to privacy and security concerns, certain investigations~\cite{shi2023towards,zhang2022game} have identified fairness issues that permeate a number of facets of federated learning, including client selection, model optimization, and contribution allocation.

To date, privacy, security, and fairness have primarily been explored in isolation. However, recent studies have revealed that there are likely to be intricate interdependencies among these constructs ~\cite{cummings2019compatibility,padala2021federated,gu2022privacy,ozdayi2021impact,furth2022fair}. For instance, researchers have found instances where the pursuing fairness can potentially compromise privacy~\cite{chen2023privacy,cummings2019compatibility,pentyala2022privfairfl,padala2021federated,gu2022privacy}, and where increasing security has impacted fairness~\cite{ozdayi2021impact,furth2022fair,xu2020reputation}. As another example, differential privacy is one of the most widely used techniques for preserving privacy  ~\cite{zhu2020more}, but scholars are beginning to discover that, beyond affecting model accuracy, differential privacy may also exacerbate issues with fairness ~\cite{cummings2019compatibility,chen2023privacy}. Moreover, when a federated learning framework becomes vulnerable to attacks, such as when a benign global model is substituted with a poisoned counterpart, it not only compromises security but also impacts fairness\cite{ozdayi2021impact}. These nascent findings have the potential to reveal fundamental interconnections between privacy, security, and fairness within the realm of federated learning.

We posit that the root cause of security and privacy concerns in federated learning lies in the practice of gradient sharing. Building upon this premise, we contend that fairness might function as a pivotal conduit between privacy and security, facilitating the attainment of heightened security and enhanced privacy within the model. Furthermore, we assert that trade-offs exist between privacy and fairness, as well as between security and fairness in the context of federated learning. Embarking on a more extensive exploration of the intricate relationships binding these dimensions should bolster our research and development efforts in federated learning to advance the field in diverse and meaningful ways.

\begin{figure}[h]
 \centering
 \includegraphics[scale=0.6]{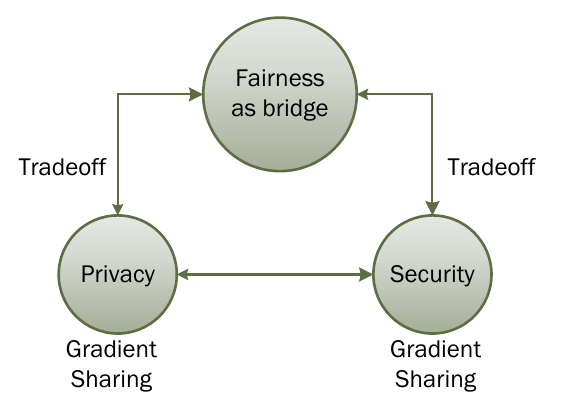}
 \caption{New Balances of Privacy, Security, and Fairness}
 \label{fig:fl_linkage}
\end{figure}

Despite the abundance of prior surveys addressing privacy, security, and fairness~\cite{abdulrahman2020survey,blanco2021achieving,gosselin2022privacy,mothukuri2021survey,truong2021privacy,zhang2022challenges,shen2022distributed}, it is notable that these surveys typically treat privacy and security as distinct entities. We posit that, for a comprehensive understanding of privacy, security, and fairness in federated learning, it is imperative to identify the underlying nexus connecting these dimensions, rather than simply addressing each concept in isolation. To this end, we have undertaken an exhaustive examination of the pertinent research conducted in recent years.  Our aim is to crystallize the fundamental interconnections between security, privacy, and fairness as it pertains to federated learning. We also intend to address existing challenges, offering fresh perspectives on achieving a nuanced equilibrium between our three constructs. The hope is that this survey will yield novel insights to help resolve privacy, security, and fairness concerns in federated learning. For example, Figure~\ref{fig:fl_linkage} illustrates that the considering fairness is pivotal when contemplating privacy preservation within the federated learning paradigm. However, it is noteworthy that the prevailing privacy technologies may inadvertently attenuate fairness, and that security issues can also create fairness concerns. Consequently, fairness emerges as a bridge that can traverse the divide between privacy and security, ultimately establishing a trade-off relationship among these three fundamental dimensions.  Moreover, at the core of the security and privacy concerns in federated learning lies the challenge of gradient sharing.

\vspace{-2mm}

In this paper, we present a thorough and inclusive exploration of the intricate interplay between privacy and security within the realm of federated learning. Our examination also extends to evaluating any methods designed to promote fairness. The methods are categorized by type and accompanied by a comprehensive analysis of the potential attacks and defense methods currently known. The survey concludes with promising avenues for future research.

The principal contributions of this survey can be summarized as follows:

\begin{enumerate}
\item[•] This survey explores the equilibrium between privacy, security, and fairness within federated learning frameworks. The intricacies associated with gradient sharing are also discussed. We find that trade-offs exist between privacy and fairness and between fairness and security, with fairness serving as a bridge in this complex interplay.

\item[•] We introduce the concept of fairness as it pertains to federated learning, describing and analyzing numerous methodologies to achieve and increase fairness. In addition, we have undertaken a comprehensive examination of privacy and security concerns, exploring them through various lenses and offering insights to shape the direction of future research. These perspectives span considerations relating to adversaries, the trained models, gradient information, and data partitioning – each of which contributes to our understanding of potential privacy breaches. In terms of security threats, we consider the integrity of the trained model, gradient data, the integrity of the training dataset, and potential adversarial actions.

\item[•] We discuss potential research directions poised to advance our understanding of federated learning and the capabilities of the field.
\end{enumerate}

The structure of the remaining article is as follows: In Section ~\ref{sec:2}, we offer an overview of federated learning  and introduce generic privacy and security techniques relevant to federated learning. Section ~\ref{sec:linkage} provides a comprehensive summary of the entanglement  between privacy, security, and fairness within the federated learning paradigm. Section ~\ref{sec:4} furnishes an essential background on fairness in federated learning. Section ~\ref{sec:tax-privacy} offers a taxonomy of privacy considerations, encompassing privacy leakage attacks, privacy-preserving methods, and their applications. In Section ~\ref{sec:security}, we provide a taxonomy of security threats, defense mechanisms, and their practical applications. Furthermore, Section ~\ref{sec:challenge} spotlights the current challenges and delineates promising avenues for future research. Finally, Section ~\ref{sec:conclusion} concludes this survey by summarizing the key findings and offering a conclusive perspective on the subject matter.

\section{Preliminary}\label{sec:2}

\subsection{Notations}
The notations used in this article are listed in Table~\ref{table1}.

\begin{table}[h]
\centering
\caption{Summary of Notations Used in the Article}
\resizebox{\linewidth}{!}{
\label{table1}
\begin{tabular}{cccc}
\hline
Notations & Description & Notations & Description\\ \hline
$M$ & A randomized algorithm & $A$ & The sensitive attributes of the data\\ 
$D$ & The datasets &$Y_p$  &The predictor outputs predicted outcome   \\ 
$D'$ & The neighboring datasets &$Y$  & The labels of the data \\ 
$\epsilon$ & The privacy budget & $i^{th}$ &  Each round of training\\ 
$\Omega$ & Every set of outcomes & $(x,y)$  &$x$ and $y$ are the input and label of the data \\ 
$H$ & Homomorphic encryption method & $\bigtriangledown {W_i}$ &The gradient of the data\\ 
$KeyGen$ & The key generation function &$(x',y')$  &$x'$ and $y'$ are the input and label of the dummy data \\ 
$Enc$ & The encryption function for asymmetric encryption &$\bigtriangledown{W^i}$ & The gradient of the dummy data \\ 
$Dec$ & The encryption function for symmetric encryption &$\bigtriangledown{W^G}$ & The difference between the dummy gradient and the shared gradient\\ 
$Eval$ & The evaluation function &$\bigtriangledown{m}$ &The malicious global model\\ 
$f$ & The query function & $\bigtriangledown {b}$ &The benign global model\\ 
$x$ & The private value &$\bigtriangledown{p}$ &The model is updated after the final poisoning\\ 
$P_i$ & The party & $c_s$ &The source class\\ 
$n$ & The number of parties &$c_m$ &The modified class\\ 
$y_i$ & The output value & $\bigtriangledown{l}$ &The locally trained model\\ 
$t$ & The number of interaction sequences &$G_i$ &The global model\\ 
$W$ & A model &$d_c$ & The small subgroup\\ 
$C$ & A client  &$L^{t}_{i+1}$ &The local model\\ 
$X$$\in$$R^n$ & The data in classification tasks &$N(0, \sigma^2)$  & The Gaussian noise\\  \\ \hline
\end{tabular}}
\end{table}

\subsection{Federated Learning}
\begin{figure}[h]
 \centering
 \includegraphics[width=0.6\textwidth]{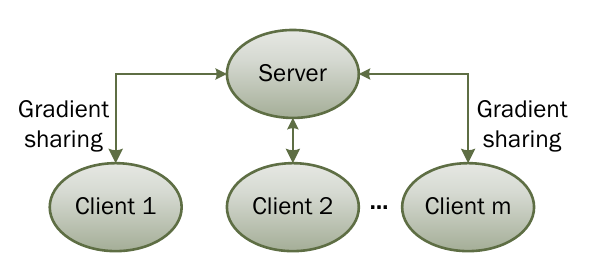}
 \caption{FL training process}
 \label{fig:fl_training}
\end{figure}

Federated learning was conceived by Google ~\cite{mcmahan2017communication}, who introduced it as an innovative departure from conventional machine learning paradigms. The fundamental objective in federated learning revolves around decentralizing training in a way that preserves privacy by training with the data distributed across local devices on those devices without ever needing to transmit those data to another site or device. Thus, the confidentiality of the underlying data sources is preserved. This stands in stark contrast to the conventional approach to machine learning, where user data is centrally stored on a server for the purposes of training a model. Federated learning strategically disseminates the model’s training across multiple clients – a paradigm shift that not only enhances efficiency but also improves the scalability of the learning process ~\cite{mothukuri2021survey}. In fact, as the volume of data continues to grow, federated learning is witnessing constant surges in popularity ~\cite{hu2016energy}. For a visual representation of the basic federated learning process, see Figure ~\ref{fig:fl_training}.

The participants in the training process are categorically divided into two distinct roles: the central server and the clients. A fundamental tenet of this framework is that the central server is deliberately precluded from accessing the local datasets of individual clients. Instead, its role is confined to receiving the client parameters only, such as gradients and weights. These parameters are subsequently aggregated to create a global model. Hence, each client retains sole possession of their own local dataset, deploying it exclusively for local training purposes~ \cite{mothukuri2021survey}.

\subsection{The privacy and security challenges in federated learning}

Federated learning offers a framework wherein multiple clients collaborate to train a global model. While federated learning mitigates the need for clients to share their raw data, sensitive information can still be revealed simply by exchanging the model parameters. In fact, recent research has demonstrated that federated learning continues to pose various both privacy and security concerns~\cite{zhu2019deep,melis2019exploiting}. Thus, several challenges to safeguarding privacy and ensuring security persist. A comprehensive outline of all these challenges follows.

\begin{enumerate}
\item[] \textbf{Model Training.} Like all machine learning models, federated learning models are susceptible to attacks, which can lead to privacy breaches and security vulnerabilities. In instances where a trained model becomes the target of an attack, there is a risk that the model updates might be manipulated before being sent to the server. For example, an adversary may seek to poison a local model with the objective of diminishing the overall accuracy of the global model~\cite{shejwalkar2022back}.  In terms of privacy, an adversary could infer whether a particular data record was used to train the global model, yet still be unable to ascertain whether a data record was used to train a specific local model~\cite{nasr2019comprehensive}. The revelation of these vulnerabilities has introduced additional and formidable security and privacy challenges to the field.

\item[] \textbf{Gradient.} Before a client shares their local gradient updates with the server, adversaries can exploit those gradients to pose privacy and security threats. For example, adversaries can surreptitiously obtain local gradients and leverage them to reconstruct a client’s private training data~\cite{zhu2019deep,zhao2020idlg}. Such unauthorized access would allow an adversary to completely recover a client’s data samples. In terms of gradient security, an adversary could poison the training data or the trained models by generating detrimental gradients that compromise the integrity of the training process. These threats represent significant challenges for federated learning.

\item[] \textbf{Training Data.} A salient feature of the federated learning training process is the complete lack of local data on the server. As such, an adversary might attempt to deduce representative classes so as to identify data samples of specific classes or even reveal the identity of a client. When an adversary poisons the training data, it not only undermines the integrity of the global model, it will also impact the performance of local models. Data poisoning attacks can be categorized into clean-label and dirty-label attacks. In a clean-label attack, the adversary cannot modify the labels of any training data, whereas in a dirty-label attack, the adversary will deliberately misclassify a substantial number of data samples~\cite{tolpegin2020data}. Additionally, an adversary might infer information about the properties of the training datasets~\cite{melis2019exploiting}, employing both passive and active techniques to deduce those properties or those of other participants involved in the training. 

\end{enumerate}
The trained models, the gradients, and the training data collectively constitute crucial components of the federated learning paradigm. Each of these elements presents a significant point of vulnerability with regard to privacy and security. The specific targets of such attacks and the defensive strategies that can be deployed to protect against them are discussed next.

\subsection{Defense methodologies}
\subsubsection{Differential privacy}
Differential privacy, initially introduced by Dwork~\cite{DBLP:conf/icalp/Dwork06}, is a stringent privacy guarantee which ensures that the manipulating one database item will have no discernible impact on the outcomes of a query. Thus, the overarching objective of differential privacy is to safeguard the privacy of a dataset's contribution by obfuscating their collective information. This privacy-preserving framework has extensive utility in the domain of federated learning, with applications spanning real-world scenarios such as electronic health data~\cite{choudhury2019differential} or artificial intelligence~\cite{zhu2020more}. 





\subsubsection{Homomorphic encryption}

Homomorphic encryption method $H$ 
\cite{paillier1999public,goldwasser1982probabilistic}is a technique that allows operations to be performed on a piece of  ciphertexts without needing to  know the decryption key. It particularly allows specific algebraic operations to be performed on encrypted content. Homomorphic encryption methods are classified into three categories \cite{acar2018survey}: partial homomorphic encryption, somewhat homomorphic encryption, and fully homomorphic encryption. 



\subsubsection{Secure multiparty computation }
Secure multi-party computation allows a party to compute a function of input values, where each party can only access its corresponding output value and cannot obtain the input and output values of other parties. For instance, if a private value $x$ is allocated to sharing among $n$ parties, all parties could cooperate in computing: 
\begin{equation}
    y_1,...y_n = f (x_1,..x_n)
\end{equation}
but each party $P_i$ could only access the content of $x_i$. Additionally, $P_i$ can only know the output value $y_i$ based on its input $x_i$, and could not obtain any additional information. 


Secret sharing is another framework for secure multi-party computing. It involves splitting secret values into random portions, and distributing them to different parties such that each party can only receives one value~\cite{shamir1979share,beimel2011secret}. Depending on the context, all or a certain number of shared values are required to reconstruct the original secret value. The main types of secret sharing ~\cite{damgaard2012multiparty}, Shamir secret sharing \cite{shamir1979share}, and binary secret sharing \cite{wang2007two}. For instance, Shamir secret sharing is based on polynomial equations. It uses matrix operation acceleration methods and satisfies theoretical information security.

\subsubsection{Knowledge distillation}
Knowledge distillation is a methodology aimed at addressing the challenge of compressing either a large model or multiple models into a more compact representation, all while preserving the performance levels of the original models. The approach hinges on a teacher-student training framework, where a trained teacher model imparts its knowledge, and a student model assimilates this knowledge through a process known as distillation training. This transfer of knowledge from a complex teacher model to a simplified student model typically incurs only a marginal performance loss~\cite{hinton2015distilling}. Various distillation techniques for strengthening privacy have been devised specifically for federated learning frameworks. 



\subsubsection{Anomaly detection}
Anomaly detection means to identify unexpected behaviors or patterns in data~\cite{chandola2009anomaly} and is a critical aspect of numerous studies~\cite{chalapathy2019deep}. Also known as abnormalities, deviants, or outliers, the core premise of anomaly detection primarily revolves around applying statistical analysis techniques to discern events that deviate from anticipated patterns or normal activity.

In the context of machine learning security defense mechanisms, approaches based on anomaly detection primarily focus on identifying  anomalous model updates and deploying outlier detection methods. These methodologies serve the overarching objective of eliminating models that deviate significantly from the norm, all the while preserving the utility of the model itself. Moreover, deploying in a range of anomaly detection methods increases the chances that  different threat scenarios will be identified in a timely fashion.

\subsection{Fairness in federated learning}\label{sec:fairness}
In the context of machine learning, fairness in the context of machine learning is frequently defined as the safeguarding of specific attributes. However, within the realm of federated learning, fairness introduces distinctive challenges, primarily attributable to the number of datasets held by individual clients and the number of classes within them. This diversity can create notable performance disparities between clients, ultimately culminating in inequitable outcomes.

Conventional fairness paradigms in machine learning typically rely on centralized datasets for both training and model evaluation. Yet directly applying these methods to federated learning does not necessarily ensure fairness across the entire population~\cite{hu2023fair}. This is because, in federated learning, fairness needs to be achieved at the individual client level, which involves minimizing any performance discrepancies between any two clients. Further, at a group level, it is imperative to ensure that the model does not discriminate against specific demographics~\cite{wang2023mitigating}. So, while most fairness methodologies rely on centralized datasets, the unique characteristics of federated learning mandate a more nuanced approach to this delicate issue~\cite{hu2023fair,wang2023mitigating}.

\begin{definition}[Fairness \cite{li2021ditto}]
If a client $C_1$'s model $W_1$  has a more uniform distribution of test performance across the network than another client $C_2$'s model $W_2$ , then $W_1$ is considered more fair than $W_2$.
\end{definition}

Notably, not every fairness constraints can be satisfied in a federated learning framework, except in very rare and special cases. Thus, researchers instead strive to achieve some type of fairness. Types include: selective participant fairness ~\cite{zhou2021loss}, which reflects the interests of both the clients and the federation. It protects system performance overall based on some probability that a client will be selected. Good-intent fairness~\cite{mohri2019agnostic} reflects the interests of the federated model via the minimum deviation of the protected attribute. Other types of fairness include contribution fairness ~\cite{lyu2020collaborative}, regret distribution fairness~\cite{yu2020fairness}, and expectation fairness~\cite{yu2020fairness}.

Equalized odds and equal opportunity can be used to measure the impact of this trade-off in classification tasks. In this task, assume the data $X$ $\in$ $R^n$ with a sensitive attribute $A$ $\in$ $\{0,1\}$ and the labels $Y$ $\in$ $\{0,1\}$. The predictor outputs the predicted outcome $Y_p$ $\in$ $\{0,1\}$. The model predicts independent results for $A$ but is accurate for $Y$.

\begin{definition}[Equalized odds \cite{hardt2016equality}]
If a predictor $Y_p$ satisfies the equalized odds criterion with respect to the protected attribute $A$ and label $Y$, and $Y_p$ and $A$ are conditionally independent given $Y$, then the predictor is said to satisfy equalized odds. 
\end{definition}
\begin{equation}
    Pr \{ Y_p = 1 | A = 0, Y = y \} -  Pr \{ Y_p = 1 | A = 1, Y = y\} = 0, y \in \{ 0,1\}
\end{equation}
The equalized odds criterion requires constraints to ensure that the true positive rates are equal to the false positive rates across all protected attribute groups, Models that perform well only on the majority group are penalized. 

\begin{definition}[Equalized opportunity \cite{hardt2016equality}]
If a predictor $Y_p$ satisfies the equalized opportunity criterion with respect to the protected attribute $A$ and label $Y$, we have the following:
\end{definition}
\begin{equation}
    Pr \{ Y_p = 1 | A = 0, Y = 1 \} -  Pr \{ Y_p = 1 | A = 1, Y = 1\} = 0
\end{equation}
In a supervised learning task, the goal of equal opportunity is to have access to the training data on the label and obtain the true label $Y$ from any equivalent prediction data. Additionally, limiting the accuracy of the model with the prediction samples ensures the fairness of sensitive attribute $A$.

Demographic parity is another form of non-discrimination. This mode of fairness requires that a decision such as  to be independent of protected attributes, such as such as accepting or rejecting a loan application will be made independent of protected attributes. The definition is as follows: 

\begin{definition}[Demographic parity \cite{gu2022privacy}]
When a predictor $Y_p$ predicts the outcomes $Y$ independently of the sensitive attribute $A$, the decision is only considered to be independent of the protected attribute if the positive prediction rate of the sensitive attribute $A$ is the same for all groups. 
\end{definition}
\begin{equation}
    Pr \{ Y_p = 1 | A = 0 \} -  Pr \{ Y_p = 1 | A = 1\} = 0
\end{equation}

Demographic parity ensures that all groups receive positive outcomes at the same rate. However, demographic parity can also be used to limit the disparity between the basic ratios of two groups.

\section{The links between PRIVACY, SECURITY AND FAIRNESS}\label{sec:linkage}

\subsection{The fundamental basis of security and privacy in federated learning}
\subsubsection{Gradient sharing}
Sharing of gradients plays a key role in determining the extent of privacy leaks and security threats within the context of federated learning. Unlike traditional distributed learning paradigms that share data, in federated learning, the gradients are shared, but this practice still leaves open vulnerabilities to privacy and security risks.

In terms of privacy, adversaries can directly manipulate gradients  to launch reconstruction attacks,  and, in so doing, potentially recover raw data from victim clients. Alternatively, they can indirectly manipulate a gradient to infer sensitive information about a client. In terms of security, adversaries have the ability to directly modify the gradient to poison the model, or they can indirectly manipulate the gradient by tampering with the data, leading to data poisoning. Such data poisoning attacks will compromise the global model's integrity, creating a significant security threats.

Shared gradients also present a formidable challenge  to privacy of federated learning frameworks, especially at the client level. Numerous studies have demonstrated that gradients can be directly used to reconstruct training data ~\cite{zhu2019deep,zhao2020idlg,geiping2020inverting,yin2021see,wang2019beyond}, and that various types of information can be indirectly inferred from analyzing shared gradients ~\cite{hitaj2017deep,melis2019exploiting,fredrikson2015model,truex2019demystifying,nasr2019comprehensive,shokri2017membership}. These reconstruction and inference attacks pose a significant threat to client-level privacy.

Clients are particularly susceptible to such attacks. This is because clients transmit their local parameter updates to the server, and, while an adversary cannot directly access the training data from the server, they can exploit the local parameters and launch a white-box gradient attack aimed at extracting the client’s private training data. Moreover, the server, while ostensibly acting in good faith, may hold a degree of curiosity. It performs parameter aggregation and shares updated global parameters with the 
clients as part of the training process. However, an inquisitive server might analyze periodic updates from clients to glean information about them. The server might even deliberately isolate certain clients and engage in an active privacy attack to acquire additional training information. Further, even if the communication between the client and the server is secure, a client under attack may still fall victim to a privacy breach before successfully uploading its local update to the server.

One notable approach to privacy attacks is the gradient-based reconstruction attack,  where the goal is to recover the victim’s training data through gradient matching~\cite{zhu2019deep}. In short, the attack works by generating dummy inputs and labels and then computing a dummy gradient based on this synthetic data. The goal is to minimize the discrepancy between the dummy gradient and the real gradient, effectively bringing the dummy data closer to the real data to facilitate the attack. Often, the attack terminates once the dummy gradient converges to the gradient of the training data~\cite{zhao2020idlg,zhu2019deep}. Alternative methods of this attack have also been explored, such as leveraging the input of a fully connected layer~\cite{geiping2020inverting,yin2021see}. Conversely, malicious clients might access the model during each training iteration~\cite{hitaj2017deep} and use generative adversarial networks (GANs) to reconstruct the private data of other clients. In essence, gradient sharing in federated learning presents a complex interplay of privacy and security challenges, necessitating vigilant defenses against a range of attack vectors.

The recovery framework proposed in many studies \cite{zhao2020idlg,zhu2019deep} can be viewed as an optimization process. In each $i^{th}$ round of training, the gradient of the data is computed where $(x,y, W_i)$, $y$ represents the data label, and $F(x,W_i)$ represents the differentiable learning model with the parameters $W_i$, $x\in R^n $ is computed:
\begin{equation}
    \bigtriangledown{W_i} = \frac{\partial(F(x,W_i), y)}{\partial{W_i}}
\end{equation}

The dummy data$(x^{'},y^{'})$ can be set randomly, where $x^{'}$ and $y^{'}$ are the input and label input of the dummy data, respectively. These dummy data are fed into the model to compute the gradient:
\begin{equation}
    \bigtriangledown{W^{'}_{i}} = \frac{\partial(F(x^{'},W_i), y^{'})}{\partial {W_i}}
\end{equation}

The goal of training is to minimize the difference between the dummy gradient and the shared gradient so that the dummy data will be close to the actual training data. This process continues until the reconstructed data converges to the actual training data.
\begin{equation}
    \bigtriangledown W_G = arg\min{||} \bigtriangledown{W} - 
     \bigtriangledown W^{'}{||}^{2}
\end{equation}

\subsubsection{Making inferences from the gradient}
In the realm of machine learning, there are many types of inference attacks, with membership inference attacks and property inference attacks being among the most common. These attacks can manifest in both black-box and white-box settings and can be further categorized into active and passive forms. In the case of membership inference attacks~\cite{hu2021membership}, machine learning models can exhibit distinct behaviors and parameters based on the training data of different clients. These discrepancies can inadvertently encode information about the training data, allowing the adversary to construct a threat model. This threat model will let the adversary differentiate between members and non-members, making membership inference attacks a central concern for systems that rely on a FedAvg algorithm. 

Moreover, membership inference attacks can directly affect the gradient, potentially triggering the target model. Such attacks may involve active tampering with the training model’s gradient ~\cite{melis2019exploiting} or observing non-zero gradients to infer the presence of specific words in the training dataset~\cite{nasr2019comprehensive}. Model poisoning attacks~\cite{sun2021fl,shejwalkar2021manipulating,shejwalkar2022back} involve malicious clients directly tampering with model updates during training, such as maliciously manipulating a gradient directly to degrade the accuracy of the global model, or doing so indirectly via a data poisoning attack~\cite{tolpegin2020data}. Here, a malicious client will contaminate the training data to contaminate the resulting gradient. However, direct manipulations of the gradient are the more potent form of attack.

Notably, poisoning attacks can occur with or without knowledge of the server’s aggregation rules. Following an attack, the global model parameters shift in the direction influenced by the attack.  because the server will aggregate both the benign and contaminated gradients to produce the global model. Here, the manipulated malicious gradient is denoted as ${\bigtriangledown}^{m}_{{i\in[m]}}$, while ${\bigtriangledown}^b$ represents the aggregation of benign gradients from clients. After the final poisoning, the model is updated to  ${\bigtriangledown}^p$. Importantly, once the global model is contaminated, its impact persists in subsequent training rounds, even without further attacks~\cite{sun2021fl}.

Additionally, clients can demonstrate malicious behavior or can fall victim to an adversary’s attacks, which might potentially involve including counterfeit or harmful samples in their local training data. Adversaries can also poison the global model by manipulating or influencing benign clients under the assumption that the server is honest and uncompromised. Data poisoning attacks mainly follow one of two key methods: label flipping~\cite{tolpegin2020data} and backdoor attacks ~\cite{wang2020attack}. In label flipping attacks, adversaries tamper with the labels of training data but do not change the data’s features. Conversely, backdoor attacks involve modifying a single feature or a small subset of the raw training data and labeling it as a specific target class. Another approach to contaminating data is by maintaining the labels but introducing malicious data. Label flipping is often used as part of a poisoning attack. Within the training data, certain data initially belonging to source class $c_s$ is altered to be classified as target class $c_m$, represented as $c_s \to c_m$. Despite modifying the data $D_i$, this also impacts the locally trained model  ${\bigtriangledown}^l$. A malicious gradient $g_m$ and is sent to the server, which is then aggregated, creating a malicious global model ${\bigtriangledown}^m$ instead of a benign global model ${\bigtriangledown}^b$.

The issues associated with sharing gradients span privacy and security attacks, all stemming from the act of sharing gradients that subsequently affects the global training scheme. Further, these attacks can involve manipulating the gradient either directly or indirectly to achieve their objectives. Addressing these gradient-sharing issues represents a significant challenge in thwarting adversary-initiated attacks within federated learning frameworks.

\subsection{Fairness as a bridge} \label{sec:fairness-bridge}

In federated learning, the relationship between privacy and security serves as a pivotal bridge that now also spans fairness considerations thanks to recent research. Within this intricate web of interdependencies, there exists several discernible trade-offs: one between privacy and fairness and another between security and fairness. Striking a balance between safeguarding the privacy of individual clients and upholding the principles of fairness has become a particularly paramount concern. Here, it is crucial to acknowledge that privacy-preserving technologies, while indispensable, can exert a negative influence on fairness. Consequently, we must meticulously scrutinize the intricate relationship between privacy and fairness.

Simultaneously, the security challenges that federated learning grapples with can also potentially give rise to issues of unfairness. This necessitates another thorough examination this time of the trade-off between security and fairness. 

The origin of fairness and bias in federated learning is multifaceted. We have known for some time that machine learning models often exhibit unexpected behaviors that have inadvertently led to some models reflecting negatively and unfairly on specific user groups. This discrimination encompasses traditional bias, disparities arising from heterogeneous data sources, selection bias in the choice of participating parties, as well as bias introduced during the aggregation of algorithms. Each of these factors contributes to the complex landscape of fairness concerns within federated learning.

\begin{enumerate}
\item[] \textbf{Traditional biases,} akin to that observed in centralized machine learning, can also be persist in federated learning. Such biases includes prejudice, underestimation, and the influence of historical negative patterns \cite{kamishima2012fairness}. In federated learning, where multiple clients participate, each entity contributes to the global model, potentially introducing or reinforcing biases during the process of updating their individual models. Additionally, the dynamics of interaction between clients and the central server throughout the training phase can further impact the fairness of the final model outcomes.

\item[] \textbf{Data heterogeneity.} Bias can also stem from the inherent heterogeneity of data in federated learning. Distinct clients possess unique data characteristics, encompassing variations in data distribution and dataset sizes. For example, if all clients used the same local batch size, those with larger datasets would require more training steps for their local models. Consequently, this might result in a substantial disparity in model outcomes. Hence, the selection of clients for each communication round can significantly influence the fairness of the final model.

\item[] \textbf{Party  selection and drop-outs.} The process of selecting which clients will participate can also introduce fairness biases. During each communication round, the server randomly chooses a subset of clients to collaborate in training.  However, there is no inherent guarantee that this selection process will faithfully mirror the true distribution of the population. Additionally, the criteria for client participation may in itself be biased. For instance, if the server repeatedly favors particular types of clients, it can lead to an  over-representation of that client group in the final aggregation model, i.e., bias.

\item[] \textbf{Algorithm biases} can also emerge due to the influence of the "majority rule". As an example, an algorithm might assign a greater weight to clients with larger datasets, potentially magnifying the impact of over or under-representing specific clients in the dataset. Consequently, when the global model undergoes training across multiple clients, those with more extensive training data tend to exhibit lower error rates compared to clients with more limited datasets, exacerbating potential fairness concerns.
\end{enumerate}

\subsection{The trade-off between privacy and fairness}
\subsubsection{The impact of privacy on fairness}
Federated learning  ostensibly prioritizes the protection of privacy, but this focus may come at the cost of some sacrifice to fairness. Homomorphic encryption and differential privacy are common techniques for addressing privacy concerns. However, the cost associated with implementing these methods is a decline in model accuracy, which, in turn, contributes to a “poor get poorer” effect, potentially exacerbating inequality.

For instance, consider a facial image model that uses differential privacy to protect gender and age classifications. An unfair model might see individuals with darker skin tones suffer greater decreases in accuracy than their lighter-skinned counterparts. Likewise, within deep models, applying differential privacy can result in varying levels of accuracy decline, with more pronounced decreases in models that were originally less accurate.

The standard definition of differential privacy, as introduced by Dwork ~\cite{DBLP:conf/icalp/Dwork06}, stipulates that a random mechanism denoted as $M : D \to R$, with a domain $D$ and range $R$, satisfies $(\epsilon, \delta)$-differential privacy if, for any two neighboring datasets $d, d' \in D$, and for any subset of outputs $S \subseteq R$, the following condition holds: $Pr[ M(d) \in S] \leq e^\epsilon \cdot Pr[M(d') \in S] + \delta$. Prior to applying of differential privacy to a specific dataset, a privacy budget must be established, where $\epsilon$ represents the privacy budget and $\delta$ signifies the probability of privacy breach in differential privacy. Each instance of differential privacy incurs a cost to $\epsilon$ depleting the privacy budget. Once spent, no further computations on the dataset can be made until the budget is replenished. Lower values of $\epsilon$ and $\delta$ correspond to higher levels of privacy. 

In a federated learning framework, where $k$ participants are collaboratively training a model, each training round $i$ sees the global server allocate the current model $G_i$ to a small subgroup $d_c$. Each participant $t \in d_c$ trains the model locally using their respective data, generating new local models denoted as $L_{i + 1}^t$. The global server then aggregates these participant-generated models, updating the global model according to a global learning rate $\eta_g$. This update process follows the formula: $G_{t + 1} = G_t + \frac{\eta_g}{k}\sum_{i \in d_c}(L_{i + 1}^t - G_t) + N(0, \sigma^{2}I)$, where $N(0, \sigma^2)$ represents Gaussian noise added to each update vector by a FedAvg algorithm. Here, introducing differential privacy restricts the influence of any one participant on the model.

\subsubsection{Balancing privacy and fariness}
To ensure both fairness and privacy in a federated learning network, one must consider a trade-off between the two. While privacy-preserving technologies such as differential privacy can maintain the accuracy of models, there is a cost to this approach: the stricter the privacy-preservation, the lower the fairness at a group level.  

\begin{table}[h]
\Large
\renewcommand{\arraystretch}{1.5}
\caption{Comparing Existing Literature with privacy and fairness}
\resizebox{\linewidth}{!}{
\begin{tabular}{c|cc|ccc|cc}
\hline
\centering
\multirow{2}{*}{\LARGE Reference} &
  \multicolumn{2}{c|}{\LARGE Privacy-Preserving} &
  \multicolumn{3}{c|}{\LARGE Fairness} &
  \multicolumn{2}{l}{\LARGE Privacy Guarantees} \\ \cline{2-8} 
 &
  \multicolumn{1}{c|}{\LARGE Training Data} &
  {\LARGE Sensitive Attribute} &
  \multicolumn{1}{c|}{\LARGE Equalized Odds} &
  \multicolumn{1}{c|}{\LARGE Equalized Opportunity} &
  {\LARGE Demographic Parity} &
  \multicolumn{1}{c|}{\LARGE DP} &
  {\LARGE MPC} \\ \hline
Cummings \cite{cummings2019compatibility}  & \multicolumn{1}{c|}{\ding{52}} & \ding{56}  & \multicolumn{1}{c|}{\ding{52}} & \multicolumn{1}{c|}{\ding{56}}  & \ding{56}  & \multicolumn{1}{c|}{\ding{52}} & \ding{56}  \\
Abay \cite{abay2020mitigating}      & \multicolumn{1}{c|}{\ding{52}} & \ding{52} & \multicolumn{1}{c|}{\ding{56}}  & \multicolumn{1}{c|}{\ding{56}}  & \ding{56}  & \multicolumn{1}{c|}{\ding{52}} & \ding{56}  \\
Du \cite{du2021fairness}       & \multicolumn{1}{c|}{\ding{56}}  & \ding{56}  & \multicolumn{1}{c|}{\ding{56}}  & \multicolumn{1}{c|}{\ding{56}}  & \ding{52} & \multicolumn{1}{c|}{\ding{52}} & \ding{56}  \\
Triastcyn \cite{triastcyn2019federated} & \multicolumn{1}{c|}{\ding{52}} & \ding{56}  & \multicolumn{1}{c|}{\ding{52}} & \multicolumn{1}{c|}{\ding{56}}  & \ding{56}  & \multicolumn{1}{c|}{\ding{52}} & \ding{56}  \\
Padala \cite{padala2021federated}    & \multicolumn{1}{c|}{\ding{52}} & \ding{52} & \multicolumn{1}{c|}{\ding{52}} & \multicolumn{1}{c|}{\ding{56}}  & \ding{52} & \multicolumn{1}{c|}{\ding{52}} & \ding{56}  \\
Rodríguez-Gálvez \cite{galvez2021enforcing}  & \multicolumn{1}{c|}{\ding{52}}  & \ding{56} & \multicolumn{1}{c|}{\ding{56}}  & \multicolumn{1}{c|}{\ding{56}}  & \ding{56}  & \multicolumn{1}{c|}{\ding{52}}  & \ding{52} \\
Pentyala \cite{pentyala2022privfairfl}  & \multicolumn{1}{c|}{\ding{56}}  & \ding{52} & \multicolumn{1}{c|}{\ding{56}}  & \multicolumn{1}{c|}{\ding{56}}  & \ding{56}  & \multicolumn{1}{c|}{\ding{52}}  & \ding{52} \\
Mozannar \cite{mozannar2020fair} & \multicolumn{1}{c|}{\ding{52}} & \ding{56}  & \multicolumn{1}{c|}{\ding{52}} & \multicolumn{1}{c|}{\ding{56}}  & \ding{56}  & \multicolumn{1}{c|}{\ding{52}} & \ding{56}  \\
Gu \cite{gu2022privacy}       & \multicolumn{1}{c|}{\ding{52}} & \ding{56}  & \multicolumn{1}{c|}{\ding{52}} & \multicolumn{1}{c|}{\ding{52}} & \ding{52} & \multicolumn{1}{c|}{\ding{52}} & \ding{56} \\ \hline
\end{tabular}}
\label{tab:my-table}
\end{table}
Cummings et al.~\cite{cummings2019compatibility} argue that achieving complete fairness within the privacy model is unattainable when transitioning from a dataset $x$ to an adjacent dataset $x'$, as there is no classifier $M$ that can be applied to dataset $x$ to maintain complete fairness. This challenge persists even if $x'$ is derived by adding or removing a sample from $x$. Consequently, these researchers proposes an approximate method of measuring fairness that incorporates discrimination coupled with an an efficient classification algorithm designed to maximize the probability of achieving privacy and approximate fairness while preserving utility. 

Pentyala et al.~\cite{pentyala2022privfairfl} also devised a privacy-preserving technique that mitigates bias as well. Their approach leverages secure multi-party computing protocols to collect and aggregate information on the label distribution and the values of sensitive attributes. Then, to rectify bias within the training set, weights are assigned to each sample based on those labels and values prior to training.

Another strategy to balance fairness and privacy is to extend the guarantee of differential privacy to both the training data and the sensitive attributes. Padala et al.~\cite{padala2021federated}, for example, developed a two-stage framework to ensure fairness while still protecting privacy. The first stage ensures fairness while maintaining accuracy. Demographic parity is considered along with equalized odds to ensure that the model’s predictions remain independent of the sensitive attributes within the dataset. A concurrent goal is to equalize the false positive and false negative rates of the model across different groups, regardless of their sensitive attributes. The second stage yields the privacy protection through local differential privacy, safeguarding both the training data and sensitive attributes. Here, Gaussian noise is introduced to protect the training data for stochastic gradient descent. This measure ensures that privacy breaches are not induced by an aggregator attack targeting the sensitive attributes. What this framework does is effectively decouple the training process into distinct stages, offering empirical insights into the tradeoffs between the fairness, privacy, and accuracy of federated learning models.

Gu et al.~\cite{gu2022privacy} explore strategies for striking a beneficial balance between privacy and fairness while also analyzing the impact of local differential privacy and global differential privacy on equity. In one of their experiments, local differential privacy is applied at the client level to observe its effects on fairness, while the server uses the DPfedAvg algorithm to establish global differential privacy. Splicing is standardized across the different groups to mitigate gradient-induced deviations. The experiment assesses group fairness through three distinct definitions of fairness as a measure of discrimination. The experiment assesses group fairness through three distinct definitions of fairness as a measure of discrimination. The results demonstrate that, given an appropriate level of noise and a fixed truncation boundary, both local and global differential privacy can deliver fairness. Notably, the stringency of the privacy-preserving mechanisms inversely affects the degree of fairness within the groups. Additionally, these researchers identify two sources of bias: the client model and imbalances in the training data. They also closely link the range of noise values to the magnitude of the noise, finding that a judicious application of local differential privacy allows for an acceptable trade-off between accuracy and privacy while still promoting fairness.

\subsection{The trade-off between security and fairness}
\subsubsection{The impact of security on fairness}
Arguably, the most common forms of security attack are poisoning or backdoor attacks, where a malicious client sends an incorrect model update to the server. These attacks can disrupt the normal convergence of the model or lead to results that do not accurately represent the true data distribution, introducing bias. Attacks on model fairness can be executed in two primary ways. The first method involves model poisoning, as seen in backdoor attacks, where the global model is altered to undermine model fairness. n the global model update process, the central server receives updates from all $n$ clients and uses a global learning rate $\eta$ to control the proportion of the joint model updated in each round. A typical expression of a joint federated learning model is~\cite{bagdasaryan2020backdoor}:

\begin{equation}
G^{t+1} = G^t + \frac{\eta}{n}\sum\limits_{i=1}^m (L^{t+1}_i - G^t). 
\end{equation}

In each round, denoted as $t$, the server randomly selects a subset of $m$ clients, referred to as $C_m$. It then distributes the current global model, represented as $G^t$, to these selected clients. Each client independently updates its local model by training it on its local dataset to result in a new model denoted as $L^{t+1}$. Subsequently, these clients send the updated model parameters, specifically $L^{t+1}_i - G^t$, back to the server.

To execute a malicious backdoor attack, the adversary will replace the legitimate global model update with a surreptitious backdoor model, referred to as $X$. They will also introduce a subset of trojan training data and a trigger into the model, altering its behavior. Importantly, this attack is covert, activating only when the trigger conditions are met, all while maintaining high accuracy to evade detection. A common implementation for the attack follows \cite{bagdasaryan2020backdoor}:

\begin{equation}
X = G^t + \frac{\eta}{n}\sum\limits_{i=1}^m (L^{t+1}_i - G^t).
\end{equation}

The second method is data poisoning which is often executed via a label flipping attack. This method allows for controlled manipulation of a dataset $D$ while maintaining high accuracy, albeit at the expense of model fairness. In this approach, each malicious client selectively alters some of the labels in the dataset $D$ with a probability denoted as $P$. These labels are modified from their original value $l$ to a target label $l_{tar}$, noting that any sensitive features $f$ can also be modified. For instance, in the context of image classification, a malicious client might change the original label of an image from 'cat' to 'dog', thereby poisoning the data. The objective of this attack is to generate a global model that fails to properly classify samples during testing. If sensitive features have been modified or if the model’s predictions are influenced by sensitive features, the model will ultimately be rendered less fair.

\subsubsection{Balancing security and fairness}
The biases observed in federated learning can often be attributed to significant variations in the distribution of local data, which, in turn, results in disparities in the model's performance. Ozdayi  and Kantarcioglu~\cite{ozdayi2021impact} demonstrated that data heterogeneity between different clients is a primary source of bias. It is also worthy noting that bias will usually grow at a much faster rate than accuracy will decline. Even when employing client update methods to adjust the aggregated learning rates in response to potential attacks, these defensive measures may themselves introduce unfairness into the model. Consequently, when pursuing model security, one must also consider fairness, striking a delicate trade-off.

Security vulnerabilities in a model can mean it fails to converge as expected or it converges on a biased model that does not represent the real data. Defensive strategies often involve filtering out models statistically outlying updates that deviate from the global aggregated average as anomalous or malicious behavior. Moreover, the global model itself may be considered malicious. However, while this approach might effectively address security concerns, it can also introduce bias. This is because, normally, updated data only represents a portion of the clients; hence, unfairness in the training model is only exacerbated.

In fact, compromising the fairness of the global model, as indicated by Furth \cite{furth2022fair}, can impact numerous local models, as any unfairness can increase demographic inequality and reduce accuracy. Vice versa, even a slight reduction in a model's accuracy can heighten the risk of demographic disparities, rendering the model unfair.

\begin{table}[h]
\Large
\renewcommand{\arraystretch}{1.5}
\caption{The literature on security and fairness. \ding{56} indicates that the paper did not include this information. }
\resizebox{\linewidth}{!}{
\centering
\label{tab:my-table4}
\begin{tabular}{ccccc}
\hline
\multicolumn{1}{c|}{\multirow{2}{*}{\LARGE Reference}} & \multicolumn{2}{c|}{\LARGE security} & \multicolumn{2}{c}{\LARGE fairness} \\ \cline{2-5} 
\multicolumn{1}{c|}{} & \multicolumn{1}{c|}{\LARGE Attack} & \multicolumn{1}{c|}{\LARGE Defense} & \multicolumn{1}{c|}{\LARGE Guarantees} & {\LARGE Metrics} \\ \hline
\multicolumn{1}{c|}{Xu \cite{xu2020reputation}} &Poisoning, Free-riders  &   \multicolumn{1}{c|}{Remove malicious adversaries} & Reputation mechanism & Clients’ real-valued contributions \\
\multicolumn{1}{c|}{Furth \cite{furth2022fair}} &Targeted backdoor attacks &\multicolumn{1}{c|}{ \ding{56}} & \ding{56} & Demographic parity  \\
\multicolumn{1}{c|}{Fraboni \cite{fraboni2021free}} &Free-rider attacks  & \multicolumn{1}{c|}{ \ding{56}} &Aggregation algorithm  & FedProx\cite{li2020federated} \\
\multicolumn{1}{c|}{Li \cite{li2021ditto}} &Poisoning attacks & \multicolumn{1}{c|}{Clipping, Krum\cite{blanchard2017machine}, K-norm } &Personalized method  &  Test accuracy variance \\
\multicolumn{1}{c|}{Singh \cite{singh2020fair}} &Poisoning attacks  & \multicolumn{1}{c|}{Anomaly detection} &Microaggregation, Gaussian mixtureModels  & False negative rate  \\
\multicolumn{1}{c|}{Song \cite{song2021reputation}} &Data Poisoning attacks & \multicolumn{1}{c|}{Reputation-based scheduling policy} &Reputation model &Reputation value \\
\multicolumn{1}{c|}{Chen \cite{chen2022dynamic}} &Poisoning attacks & \multicolumn{1}{c|}{Mutual evaluation mechanism } &Mutual evaluation mechanism &Correlation coefficient \\ \hline
\multicolumn{1}{l}{} &  \multicolumn{1}{l}{} & \multicolumn{1}{l}{} & \multicolumn{1}{l}{} & \multicolumn{1}{l}{} \\
\multicolumn{1}{l}{} & \multicolumn{1}{l}{} & \multicolumn{1}{l}{} & \multicolumn{1}{l}{} & \multicolumn{1}{l}{}
\end{tabular}}
\end{table}
Several methods have been proposed to strike a balance between security and fairness in federated learning. First consider that unfairness in federated learning has been attributed to two main factors: data heterogeneity and security attacks. Both have been explored in the literature.

For example, Xu and Lyu~\cite{xu2020reputation} introduced a federated learning framework that addresses both fairness and security concerns that specifically focuses on targeted poisoning attacks (e.g., label-flipping), untargeted poisoning attacks, and free-riders attacks. Their approach incorporates a reputation mechanism, where a client's reputation is determined based on a contribution value that is assessed using the test accuracy from locally trained models and gradient uploads. This helps to identify and reward contributing clients while identifying and removing non-contributors or malicious actors. However, while effective against security attacks like backdoor attacks, the security and robustness mechanisms in a federated learning framework may inadvertently filter out users with simpler data, potentially undermining fairness.

Data heterogeneity can further complicate fairness concerns, as benign clients might exhibit the characteristics of overfitting and so resemble a compromised client. This resemblance may allow backdoor attackers to disguise themselves and evade feature checks ~\cite{zawad2021curse}. Notably, data heterogeneity can amplify the weight differences between benign clients. Consequently, benign clients may become less distinguishable from malicious ones. For this reason, cosine similarity is often used to detect weight exceptions~\cite{bagdasaryan2020backdoor}. To tackle this challenge, Li et al.~\cite{li2021ditto} introduced Ditto, a personalized federated learning framework primarily designed to address data heterogeneity. When malicious attackers disrupt the process of training the global model, sending that global model to different clients will result in suboptimal outcomes. However, local benign clients may struggle to train an effective model using only their local data. Hence, Ditto offers a trade-off between personalized and global models through a hyperparameter $\lambda$. A higher $\lambda$ aligns the personalized model closer to the global model, while a lower $\lambda$ allows greater deviation from the poisoned global model. By adjusting $\lambda$, each client can find a personalized model that strikes a good balance between the global model and their own local model. Moreover, Ditto significantly improves both model accuracy and fairness under various attacks. It also minimizes the variance in test error rates such that benign clients are treated fairly. Lastly, experiments show that Ditto outperforms three other defense mechanisms in terms of global model test accuracy.

\section{Fairness }\label{sec:4}
In the previous section~\ref{sec:linkage}, we explored the intricate relationship between privacy, security, and fairness. While these concepts may initially seem distinct, they are closely intertwined in a range of ways. A significant connection exists between privacy and security that primarily revolves around the sharing of gradients, and it is this connection that can potentially give rise to privacy and security vulnerabilities. Furthermore, privacy and security both share a connection with fairness, as a trade-off relationship exists between all three factors. Hence, if the objective of a training task is to safeguard client privacy and security, a singular focus on either one will likely compromise fairness. For this reason, these objectives should not be viewed in isolation, but rather all need to be considered together. 

Having previously introduced fairness in Section~\ref{sec:fairness}  and fairness bias in Section~\ref{sec:fairness-bridge}, this section will focus on several contemporary approaches to bolstering fairness in federated learning, as well as how these approaches can be integrated into privacy and security considerations.

\subsection{Fairness enhancing methods}
\begin{figure}[htbp]
 \centering
 \graphicspath{{Federated Learning Framework/}}
 \includegraphics[width=0.6\textwidth]{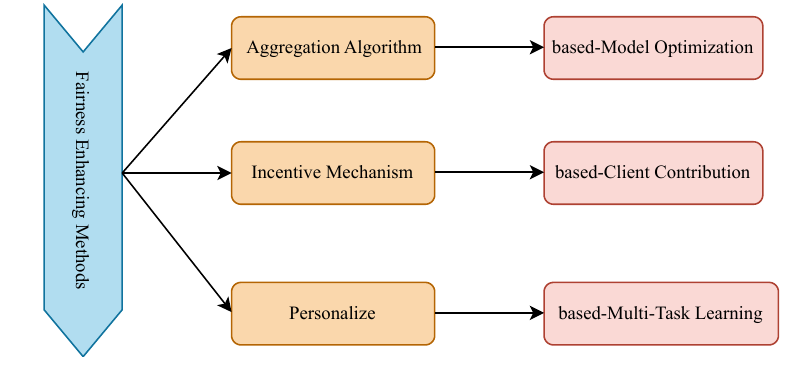}
 \caption{Fairness enhance in federated learning from different perspectives}
 \label{fig:fairness_enhance}
\end{figure}

The kind of biases and unfairness commonly encountered in centralized machine learning is also prevalent in federated learning. Extensive research has been undertaken to tackle this issue, typically falling into three primary categories: (1) enhancing the aggregation algorithms; (2) establishing incentive mechanisms; and (3) tailoring the methodology, as illustrated in Figure~\ref{fig:fairness_enhance}. 

\textbf{Type 1, Aggregation Algorithm}. 
To combat bias and unfairness in federated learning, numerous studies have proposed enhanced aggregation algorithms~\cite{wang2020federated,li2020federated,li2019fair, mohri2019agnostic,hu2020fedmgda+,huang2020fairness}. For example, both Li et al.~\cite{li2019fair} and Mohri et al.~\cite{mohri2019agnostic} have published novel optimization objectives to address fairness concerns. Mohri et al.~\cite{mohri2019agnostic} proposed the concept of benevolent fairness to safeguard the worst performance of a random client, while Li et al.~\cite{li2019fair} uses the uniformity of performance as a measure of fairness. Both studies leverage the idea of fair resource allocation to maintain overall performance and fairness.

Drawing inspiration from equitable resource allocation in wireless networks, Hu et al.~\cite{hu2020fedmgda+} introduced the parameter $q$ to represent an aggregated weighted loss, assigning a higher relative weight to clients with greater losses. Additionally, they devised methods to preserve accuracy and defend against malicious clients without compromising either accuracy or fairness. Similarly, Huang et al.~\cite{huang2020fairness} also employed a weighting strategy to promote fairness coupled with a double momentum gradient method. In the momentum gradient descent method, the exponential weighted average of the gradients is computed and then used to update the weights. The server then aggregates the weights of all clients based on the training accuracy and the number of times a client has participated in training. Thus, a change factor underpins the methodology behind this metric to align more closely with real-world scenarios.

Another solution, FedProx~\cite{li2020federated}, addresses issues with heterogeneity, showing promise in combatting bias in heterogeneous networks. Given that the client devices in most federated networks are many and varied, expecting uniform workloads from each device is neither practical nor ideal. FedProx, which is a straightforward modification of the original FedAvg algorithm, achieves superior performance while also accommodating disparities in performance. It considers factors like differences in computing power and other variables related to the equipment involved in training rounds. FedProx also introduces a regularization term to address non-uniformity in local updates.

Wang et al.~\cite{wang2021federated} takes a a different approach, framing fairness as a conflict between clients. Their framework uses cosine similarity to detect gradient conflicts, which are resolved by adjusting the direction and magnitude of the gradient prior to gradient averaging.

However, it is worth noting that the majority of studies taking the aggregation algorithm approach assume naive federated learning without and do not considering attacks. Yet, in practical situations, federated learning structures will be susceptible to these nefarus activities. Hence, fairness, privacy, and security all need to be considered comprehensively.

\textbf{Type 2, Incentive Mechanism}. In addition to algorithmic approaches, incentive mechanisms have been proposed as a means to enhance fairness and accuracy in federated learning ~\cite{michieli2021all,wang2019measure,yu2020fairness,lyu2020towards}. Yu et al.~\cite{yu2020fairness} introduces three equity criteria: contribution equity, expected loss distribution equity, and expected equity. The scheme is interesting but estimating the cost incurred by clients and quantifying their contributions is challenging.

Lyu et al.~\cite{lyu2020towards} devised a local credit rating system to ensure fairness, assessing a client's contribution to the learning process. This iterative system allows clients to converge towards a native model that aligns with their contributions while preserving client privacy through a three-tier onion encryption mechanism.

Michieli and Ozay~\cite{michieli2021all} introduced a novel aggregation approach, wherein each client's contribution to the aggregation model serves as a guarantee for equitable resource allocation. This method yields improved convergence speed and accuracy.

Wang et al.~\cite{wang2019measure} proposed a technique for fairly assessing each client's contributions. In a vertical federated learning framework, they use a Shapley value to gauge the importance of each client and determine their contribution. In horizontal federated learning, the contribution is calculated by removing an instance and retraining the model. The difference between the results of the new model and the original serves as a measure of the contribution.

In this approach to addressing fairness concerns, incentive mechanisms to motivate clients to deliver high-quality contributions to the learning program. The central challenge, though, is how to accurately compute the cost of participating in federated learning and how to estimate the client's contribution.

\textbf{Type 3, Personalize}. Another avenue for enhancing fairness in federated learning is personalized federated learning, where individual clients learn from distinct models tailored to accommodate the inherent heterogeneity in this learning paradigm ~\cite{divi2021new,li2021ditto,balakrishnan2021resource}. In practical terms, heterogeneity represents a critical source of unfairness, stemming from factors such as computational disparities, variable client arrival times at the server, communication discrepancies, and varying data transmission rates, all of which can impact the overall durations of training on a client. Furthermore, statistical heterogeneity can manifest due to  imbalanced data and differences in data distribution across clients. 

Both Divi et al.~\cite{divi2021new} and Li et al.~\cite{li2021ditto} have proposed personalized federated learning approaches to address such fairness concerns. Divi~\cite{divi2021new}, for example, established a fairness index that uantifies the extent to which a personalized model offers equitable opportunities. Balakrishnan~\cite{balakrishnan2021resource} also put forward a personalized method for ensuring fairness that focuses on the local models. Their method leverages importance sampling to tackle communication heterogeneity by harnessing client computing and communication resources. Furthermore, to overcome any issues arising from distinct data distributions, the framework personalizes the resources allocated to modelling so as to ensure improved performance for each client.

This approach to fairness empowers personalized methods to adapt to the diversity inherent in different models and federated settings for each client. Nevertheless, striking a balance between privacy preservation, model accuracy, and equitable resource allocation through a personalized approach remains a formidable challenge.

\subsection{Enhancing fairness while balancing privacy and security }
The above-mentioned methods for enhancing fairness primarily assume a fairly simplistic federated learning environment. As such, these approaches often overlook the privacy and security concerns inherent to federated learning. Thus, when attempting to ensure fairness, privacy, and security all at the same time, relying solely on a fairness algorithm will not be sufficient. This is because federated environments are susceptible to privacy breaches and security attacks by malicious participants, soo extra defenses will be required. That said, in addressing the challenge of simultaneously bolstering fairness, privacy, and security in federated learning, numerous studies have made notable progress 
~\cite{pentyala2022privfairfl,zhang2020fairfl,lyu2020towards}.

Some have posed the aim of balancing privacy, security and fairness as additional constraints in a training task, effectively turning federated learning into multiple-objective optimization problem.
Pentyala et al. ~\cite{pentyala2022privfairfl}, for instance, proposed a privacy-preserving training approach that comprises pre-processing and post-processing phases. The pre-processing phase tackles issues related to heterogeneous data distributions and assigns weights to samples, effectively mitigating bias in a unified training dataset. Additionally, the client weights are encrypted to safeguard sensitive information. This method successfully aligns the optimization objectives of privacy and fairness, offering both model privacy and group fairness. Likewise, Zhang et al. ~\cite{zhang2020fairfl} also explored the optimization objectives of privacy and fairness constraints.  They devised a Markov game framework to select to participate in each communication round, with client decisions contingent on the state of the global model. This approach maximizes fairness and accuracy while strictly preserving privacy through a security aggregation protocol that restricts each client's access to local data. Singh et al. ~\cite{singh2023fair} introduced security and fairness as optimization objectives. The goal with the security optimization is to detect any model or data poisoning activity, while the aim of the fairness optimization is to reduce client discrimination. The proposed framework considers scenarios where the mechanisms to defend against model poisoning might mistakenly filter out benign clients due to data heterogeneity. Three methods are outlined: micro-aggregation to distinguish minority members from attackers; a Gaussian mixture model to describe the distribution of client updates; and density-based clustering to characterize non-iid client update distributions.

Alternatively, several studies have integrated incentive mechanisms to bolster client fairness – the idea being that the final model distributed to clients is based on their contribution to building the model, with greater rewards allocated to clients who have made more substantial contributions. Lyu et al.~\cite{lyu2020towards} devised a mechanism for evaluating local credibility. The mechanism improves fairness while still ensuring privacy through an encryption scheme. Participants earn points for downloading gradients from other participants, with additional points attainable through sample or gradient uploads. These transactions are immutable and recorded on a blockchain, and homomorphic encryption safeguards all information during gradient uploads and downloads. Ruckel et al.~\cite{ruckel2022fairness} proposed an approach where fairness hinges on calculating the global performance of each client according to their actual parameters. 
Rewards are then assigned accordingly. To mitigate information leaks, each client’s model update is perturbed via local differential privacy.  Xu and Lyu~\cite{xu2020reputation} introduced a reputation mechanism to address security issues and enhance collaborative fairness. The mechanism scrutinizes the gradients uploaded by clients to identify and remove any malicious actors, thereby upholding security. The reputation system additionally quantifies the participants’ contributions, enabling tailored rewards based on distinct patterns of performance.

\section{Privacy}\label{sec:tax-privacy}
As mentioned, the main point of privacy vulnerabilities in a federated learning system is when gradients and model parameters ae exchanged between the clients and the central server. In this section, we aim to categorize privacy attacks and the methods used to defend against them. The section concludes with a discussion on some practical applications for safeguarding user privacy.

\subsection{Privacy attacks}

\begin{figure}[h]
 \centering
 \includegraphics[width=0.7\textwidth]{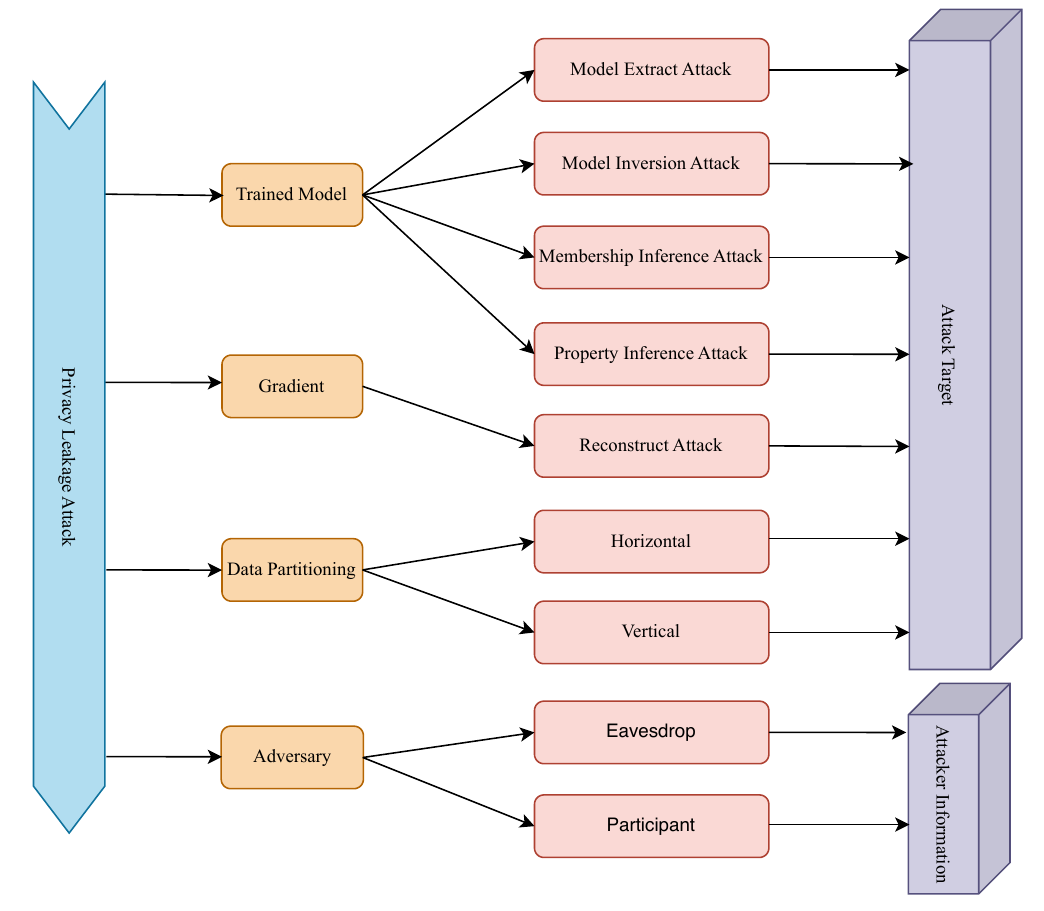}
 \caption{Privacy attacks from different multiple perspectives}
 \label{fig:privacy_leakage}
\end{figure}

By their very nature, federated learning frameworks are designed to safeguard the privacy of their participants by circumventing the need to share raw data and sharing only model updates instead. Nevertheless, recent research has revealed that vulnerabilities still exist within these frameworks that can lead to significant privacy breaches ~\cite{zhu2019deep}. In this section, we present a comprehensive classification of privacy attacks and potential privacy leaks in federated learning landscapes. The four perspectives from which we view this subject include: attacks and leaks that can occur while training a model; attacks on gradients, which are particularly susceptible to breaches; privacy from the adversary’s perspective and their access to information; and data partitioning, both horizontal and vertical. Figure~\ref{fig:privacy_leakage} illustrates these four perspectives.

\begin{table}[]
\caption{The link between privacy target and fairness}
\begin{tabular}{llll}
\hline
Privacy Target &         & Fairness & Remark                                           \\ \hline 
Model          & Attack  & Yes   & Aggravates unfairness for the clients\\
               & Defense & Yes         &The cost of protecting the model privacy means sacrificing fairness \\
Gradient       & Attack  & Yes        & Aggravates unfairness for the clients\\
               & Defense & Yes        & Adding differential privacy to the  gradients increases unfairness
  \\
Training Data  & Attack  & Yes        & Aggravates unfairness for the clients     \\
               & Defense & Yes         &Improving fairness for clients decreases the model’s fairness      \\ \hline
     
\end{tabular}
\end{table}

\subsubsection{Attacks on trained models.}
In a privacy leak, an unauthorized observer is able to see or infer all or some of the features of a target training sample simply by analyzing the output of the trained model. 

Thus, trained models sit at the nexus of privacy leaks in federated learning. In fact, there are four main types of privacy attack that target trained models: (1) model inversion attacks; (2) model extraction attacks; (3) membership inference attacks; and (4) property inference attacks.

\textbf{Model inversion attacks}.
Model inversion attacks exploit access to a model to deduce information about the training data. More specifically, the intention is to extract sensitive features about the model’s input based on some corresponding output from the model. 

The pioneering model inversion attack was introduced in the context of genomic privacy, with Fredrikson et al.~\cite{fredrikson2014privacy}demonstrating that information could be recovered from a training set even with only black-box access to the prediction model. This attack was subsequently extended to novel settings where sensitive features could be inferred from the inputs to a decision tree model ~\cite{fredrikson2015model}. These attacks are most effective in scenarios where the inferred representative class originates from a small training set. 

In cases involving malicious clients, Wang et al. ~\cite{wang2019beyond} presented a multi-task framework based on a GAN designed to construct user-level private information from input samples, such as the true category of an instance or the identity of the client. Zhang et al.~\cite{zhang2020secret} also turned to GANs to guide an inversion process, establishing a fundamental link between a model’s predictive capabilities and its susceptibility to inversion attacks.

Notably, the abovementioned methods are passive attacks. But model owners should take caution because adversaries can also actively interfere with a model’s training process. For example, Hitaj et al.
~\cite{hitaj2017deep} outline a novel class of active attacks that involve launching a model inversion attack on some GANs to reconstruct facial images of the users through white-box mechanisms. Here, the adversary uses deception to extract accurate yet sensitive information about the victim.

\textbf{Model extract attacks}. In a model extraction attacks, the adversary has no prior knowledge of the target model’s parameters or training data and so attempts to obtain the model's parameters by extracting the target model. As a result, the adversary actually increases the success rate of subsequent attacks on the model's training data.  Ateniese and colleagues~\cite{ateniese2015hacking} were the first to address the notion of extracting unexpected but useful information from a trained model. They designed a meta-classifier and trained it to hack other classifiers so as to infer sensitive information about the training set. In \cite{tramer2016stealing}, Tramèr et al. demonstrated an equation-solving method of model extraction, extending Ateniese et al.’s work. They also demonstrated successful model extraction attacks against a variety of model types that output only class labels.

\textbf{Membership Inference Attacks}.
In a membership inference attack, the goal is to determine whether a sample record has been used to train a model, i.e., to infer whether that record is a member of the training set. The consequences of such attacks can be very serious for the individuals involved. For instance, if a data record is known to have been used in a model trained to classify types of cancer, a membership inference attack could potentially leak information about the health of that individual~\cite{shokri2017membership}. Several recent studies have demonstrated that machine learning models, no matter the learning schema, are vulnerable to membership inference attacks ~\cite{hu2021membership,shokri2017membership,nasr2019comprehensive,truex2019demystifying}, even in black-box settings ~\cite{truex2019demystifying}. For example, an adversary might discern whether a data record forms part of the model’s training set through an API. In this category of attack, Nasr ~\cite{nasr2019comprehensive} devised a white-box membership inference strike that can retrieve private data from the model but fails to obtain data from a model but cannot obtain data from other models with the same distribution.

\textbf{Property Inference Attacks}.
Property inference involves identifying properties that hold for specific subsets of the training data but are not universally applicable to all class members. In these attacks, adversaries access trained models to extract global statistics about the training data ~\cite{cryptoeprint:2021/099}. In Melis et al.’s~\cite{melis2019exploiting} attack, for example, the aim is to infer all the properties of a subset of the training data, focusing on the properties independent of class-specific features. For instance, they inferred whether certain images depict individuals wearing glasses using a gender classifier. These attacks can be either passive or active. In a passive attack, the adversary aims to infer two sets of data-one with and one without attributes. These data points must belong to the same class as the target participant but may not be dissimilar in other respects. In an active attack, the adversary actively seeks to infer attributes via multi-task learning.

\subsubsection{Attacks on gradients}
While gradient exchange was once considered a secure method for uploading data to a server, recent research  has exposed  this practice as being vulnerable to attack ~\cite{zhu2019deep,zhao2020idlg,geiping2020inverting,yin2021see}. Even though the central server avoids direct access to user data, gradient attacks have demonstrated the potential to recover a user's local training data from exchanged parameter gradients. When clients share their models with a central server, malicious attackers can intercept those shared gradients and reconstruct sensitive information about the participant, resulting in privacy breaches.

Consider, for example, a case where the adversary is actually a participant in the training scheme. Under these circumstances, the adversary could optimize dummy gradients to approximate the shared gradients and, in the process, infer both inputs and labels ~\cite{zhu2019deep,zhao2020idlg}. Notably, this method tends to be less effective in deep networks. But a simpler, yet effective, approach involves extracting the ground-truth label based on the sign of the gradient vector~\cite{zhao2020idlg}. This approach is applicable to any differentiable model trained with cross-entropy loss over one-hot labels, which is the typical scenario for classification tasks.

Geiping et al.~\cite{geiping2020inverting} go so far as to  demonstrated that it is possible to reconstruct the input to a fully-connected layer from parameter gradients, regardless of the layer's position in a neural network. Their study shows that gradient inversion can faithfully reconstruct samples, even for trained and untrained parameters in deep and non-smooth architectures. Another method can recover a single image from a batch by inverting the averaged gradients and using a fully-connected layer’s gradients to recover ground-truth labels for the labels within the batch ~\cite{yin2021see}. Importantly, all these methods are capable of recovering training data without the need to transform any gradients. 

In another vein, Li et al.~\cite{li2022auditing} outlines a gradient attack method called generative gradient leakage (GGL). GGLs rely on a generative model trained on public datasets as prior information to ensure a quality reconstruction given noise and defensive transformations. To reduce the search space and enhance the quality of the generated image, the hidden expression closest to the gradient of the real image is identified within the potential space of a GAN.

In these research endeavors, adversaries approximate and reconstruct the local models of clients using gradients, which leads to privacy breaches. Consequently, safeguarding gradients has become a significant challenge in federated learning, as to at least some extent, preventing gradient leaks can mitigate privacy risks.

\subsubsection{The adversary's perspective}
\begin{figure}[!ht]
\centering
\subfloat[Eavesdropper]{
		\includegraphics[scale=0.5]{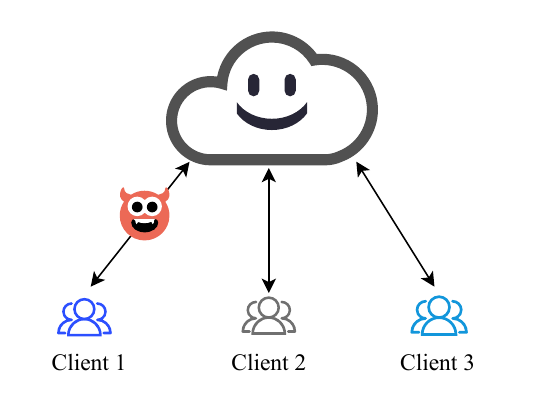}}
\subfloat[Malicious client]{
		\includegraphics[scale=0.5]{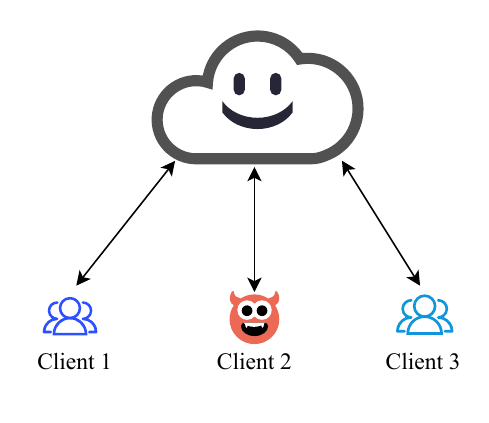}}
\subfloat[Malicious server]{
		\includegraphics[scale=0.5]{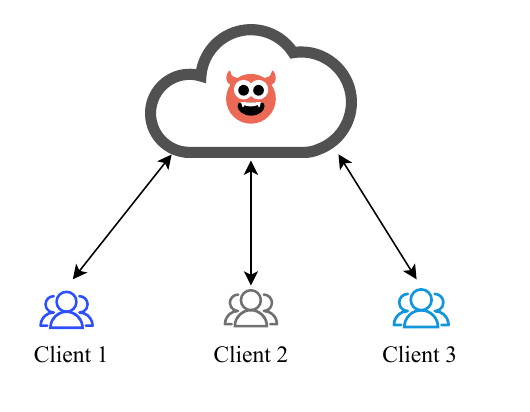}}
\caption{Adversary information in federated learning privacy attacks}
\label{fig:privacy_ad}
\end{figure}

Training in federated learning involves rounds of communication that adversaries can exploit to reveal data. Within this process, the adversary can either be an eavesdropper or a participant. Eavesdroppers observe the process without impacting performance – their objective being to expose and gather sensitive information, such as the training data or the model parameters. Participants, on the other hand, are those directly involved in the process, i.e., a client or even the central server. Hence, participants have the ability to both observe and modify the training data or model parameters. Figure~\ref{fig:privacy_ad} illustrates these three types of adversary information.

When the central server acts as an adversary, it can either be semi-honest or malicious. In a semi-honest setting, the objective is to reveal user data ~\cite{geiping2020inverting}. Here, the adversary does not typically alter the model architecture to enhance their attack, nor do they transmit malicious global parameters that do not genuinely represent the global model. However, they can access the current global model and the shared gradients~\cite{li2022auditing}. From this information, they should be able to infer sensitive information about the clients’ data~\cite{fereidooni2021safelearn}.

In a malicious setting, the adversary's goal is to reconstruct the private data of a target client, enabling client-level privacy attacks~\cite{wang2019beyond}. Attacks like gradient disaggregation fall into this category~\cite{lam2021gradient}, where a malicious central server performs the attack, potentially conducting inference attacks on the model updates of individual participants.

However, when an adversary impersonates a client, the attack’s potency increases significantly. For example,  Melis et al.~\cite{melis2019exploiting} analyzed periodic training updates and was able to infer the properties of the participants’ training data from them. However, notably, this analysis was independent of the class characteristics. By contrast, Fu et al.’s~\cite{fu2022label} approach was somewhat more novel. After completing the training process, each client receives a trained bottom model. The adversary can then train a model to infer information based on the trained bottom model and can infer labels for any sample of interest, not limited to those in the training dataset. The adversary might even take the form of a curious client~\cite{nasr2019comprehensive}, actively influencing the target model to extract more information about its training set during the training process.

When the adversary is an eavesdropper, the aim is usually to pilfer sensitive information exchanged between the clients and server, such as a training update or the final global model~\cite{elgabli2021harnessing}. Recent studies ~\cite{zhu2019deep,zhao2020idlg} reveal that eavesdroppers first create dummy data to generate a gradient via local training and subsequently use this gradient to approximate the real gradient, leading to profound privacy breaches as the original training samples can be reconstructed from this shared gradient.

\subsubsection{Data partitioning}
There are two data partitioning schemes in federated learning: horizontal and vertical. Each is contingent on how the data is distributed across the sample and feature spaces. In horizontal data partitioning, the client datasets share the same feature spaces but have different sample spaces. Conversely, in vertical data partitioning, the client datasets have different feature spaces but the same sample spaces.

In the realm of horizontal data partitioning, Google introduced an application for waking an Android phone using wake-word recognition~\cite{mcmahan2016federated}. However, Zhu et al.~\cite{zhu2019deep} revealed the potential for malicious attackers to reveal the raw features and labels in some client data simply by having knowledge of the model's architecture, parameters, and communicated gradient loss. Zhao et al.~\cite{zhao2020idlg} further demonstrated that the ground truth label of an example could be extracted by exploiting the index of the output associated with a negative gradient. Hitaj et al.~\cite{hitaj2017deep} explored how an honest client could inadvertently expose their private data to an inference attack – that is, by using GANs to generate samples intended to mimic the same distribution as the training data. Generally, the aim of these membership inference attacks is to acquire information by determining whether a sample distribution exists within a training set~\cite{hu2021membership}.

In the context of vertical data partitioning, Hardy et al.~\cite{hardy2017private} introduced a protocol tailored for linear models. However, privacy leaks within vertical data partitioning scenarios remain relatively underexplored. 

Li et al.~\cite{li2021label} argued that stringent measures are necessary to safeguard the privacy of the labels in participant data. This caution arises from a recognition that raw labels may contain highly sensitive information. On this note, Fu et al.~\cite{fu2022label} presented a novel set of label inference attacks specifically designed for vertical federated learning. In these attacks, a participant without access to the labels assumes the role of the adversary, seeking to infer the labels. Conversely, Cheng~\cite{cheng2021secureboost} proposed an end-to-end privacy-preserving tree-boosting framework, which ensures that all participants are shielded from leaking data information to one another.

\subsection{Methods of privacy preservation }
In light of the diverse methods for perpetrating a privacy attack within a federated learning setting, it is abundantly clear that safeguarding against such breaches is of the utmost significance. Fortunately, there are also numerous methods of bolstering privacy protection.

An intricate relationship exists between privacy breaches and defense mechanisms. For instance, the incorporating differential privacy into local data can be a very effective measure. This strategic addition of noise ensures that, even if an adversary does manage to recover the data through a reconstruction attack, that data will be too noisy to decipher. Similarly, when contending with a model inversion attack, introducing differential privacy to the results after the local/global model has been trained should be an effective countermeasure to stop the adversary from inferring any attributes.

Further, a substantial connection exists between confidential computation technologies and defensive measures. For example, techniques like model updates and data encryption act as formidable barriers that can stop adversaries from accessing sensitive information. Consequently, these measures effectively stymie an array of attacks, including those centered around reconstructing or inferring data from gradients.

Another defense mechanism is secure aggregation. This is a communications measure that streamlines model aggregation. It deftly strikes a balance between privacy preservation and communication efficiency, thereby ensuring data protection while enabling seamless collaboration.

Last on the list is knowledge distillation. This process circumvents the exchange of model parameters, preemptively thwarting attacks that seek to exploit these parameters so as to infer member information or recover the original data.

\subsubsection{Differential privacy}
In federated learning, collaboratively training a shared model is an iterative process involving multiple devices. Beyond not sharing private data, two of the main vulnerabilities that remain in this privacy-preserving learning paradigm are inference~\cite{shokri2017membership,melis2019exploiting} and reconstruction ~\cite{zhu2019deep,zhao2020idlg}.

Differential privacy and its two modalities – local and global differential privacy – can be a powerful safeguard against both these issues.

Global differential privacy depends on a trusted server that is responsible for injecting noise into the output, typically done during the global model updates. In Geyer et al.’s~\cite{geyer2017differentially} mechanism, the server randomly selects which participants will train a local model. This is followed by global model updates during each communication round as usual, but random Gaussian noise is injected into the global model during aggregation updates, as outlined by McMahan et al.~\cite{mcmahan2017learning}. In fact, 
McMahan et al. pioneered the application of user-level privacy protection for language models, underpinned by global differential privacy. Notably, this marked the initial application of differential privacy to averaging algorithms in federated learning for safeguarding user-level sensitive information.

Another notable contribution was by Wei et al.~\cite{wei2020federated}, who introduced the NbAFL framework. In NbAFL, artificial noise is thoughtfully infused into the participants’ parameters before any updates transpire. This proactive integration of global differential privacy by introducing only a limited amount of noise can extend the protective purview of the privacy guarantee to entire datasets, concurrently guarding the participants’ privacy while maintaining commendable levels of accuracy. Nonetheless, it is imperative to acknowledge that the Achilles’ heel of global differential privacy resides in the challenge of determining an appropriate sensitivity level, as this parameter has a substantial influence on both the level of privacy assured and the model’s performance.

Local differential privacy stands apart from global differential privacy in that it directly safeguards sensitive data. Here, noise is added locally and so there is no need for a trusted server to intervene. This noise injection can happen in one of two ways:

\begin{enumerate}
\item[•] noise is added to the user's local gradients before the parameter updates  is transmitted; or
\item[•] noise is added to the local model.
\end{enumerate}

To fortify  a system’s defenses against inference attacks, Zhao et al.~\cite{zhao2020local} introduced an algorithm known as LDP-FedSGD, tailor-made for IoT. This algorithm prevents an adversary from inferring any original data by perturbing the gradients before any updates are sent. In this setup, the server orchestrates updates to the global model by averaging the outcomes of these perturbed gradients.

Truex et al.~\cite{truex2020ldp} introduced a scheme, called LDP-Fed, that introduces perturbations to each participant's gradient based on a localized instance of  local differential privacy module. This approach diligently shields the participants' data against inference attacks.

Also tackling the realm of IoT, Cao et al.~\cite{cao2020ifed} explored an innovative framework called IFed. IFed introduces standard normal noise to users' local models to create new models that subsequently update the global model. This strategic noise injection serves as a deterrent against anyone trying to reconstruct the raw data.

Wu et al.~\cite{wu2021incentivizing} contrived an incentive-based framework that helps to protect sensitive information by infusing artificial Gaussian noise into the local model. This measure strategically thwarts any attempts at data reconstruction. Bhowmick et al.~\cite{bhowmick2018protection} also leverage local differential privacy to fortify defenses against data reconstruction during client-side communications while upholding the privacy of any model during global updates. In this way, local differential privacy emerges as a robust solution for averting privacy breaches from a participant’s vantage point, affording a more potent privacy guarantee.

In summary, within the landscape of federated learning, both global differential privacy and local differential privacy emerge as potent tools for privacy preservation. While global differential privacy finds applicability across entire datasets, it depends on a trusted server and carries the challenge of determining the correct sensitivity levels. By contrast, local differential privacy offers participants a more robust privacy guarantee.

\subsubsection{Confidential computation technology}
Confidential computation technology serves as a pivotal asset in preventing privacy breaches and enhancing accuracy within the sphere of federated learning. Federated learning faces an array of privacy threats, notably inference attacks and gradient attacks, which can be effectively mitigated through the use of three distinct categories of technology: (1) homomorphic encryption, (2) secret sharing, and (3) secure multi-party computation.

\textbf{Homomorphic encryption:} With homomorphic encryption, one can directly encrypt data without the need for a secret key. However, even with homomorphic encryption, an adversary might recover some training samples or glean membership information by inferring data from shared gradients. To counteract this concern, Liu et al.~\cite{liu2021privacy} introduced an enhanced privacy framework couplied with a public key that involves a homomorphic encryption mechanism to encrypt gradients. Similarly, Xu et al.~\cite{xu2020privacy} devised a privacy-preserving strategy for federated training processes that integrates additive homomorphic cryptosystems to guarantee the confidentiality of sensitive user information. This multifaceted approach involves initializing encrypted aggregated values, user reliability updates, aggregated value updates, and weighted federated parameter updates. Thus, homomorphic encryption not only significantly bolsters accuracy~\cite{truex2019hybrid} but can also extends formal privacy assurances by encrypting both the participants' data and the aggregated updates.

\textbf{Secure multi-party computation:} Secure multi-party computation (SMC)~\cite{canetti1996adaptively} plays a pivotal role in addressing the inherent challenge of determining whether to trust a scheme’s participants, each of whom possesses confidential data. This is because secure multi-party computation guarantees that each participant can access accurate computation results based on their own data but not any information beyond the results themselves. Bonawitz~\cite{bonawitz2017practical} devised a protocol based on secure multi-party computation to compute the sum of the aggregated updates. The aim was to not only advance techniques for preserving privacy but also enhance communication efficiency. Recent developments have witnessed the design of secure protocols intended to achieve secure predictions for deep neural networks (DNNs). In this context, Agrawal et al.~\cite{agrawal2019quotient} introduced a novel approach to secure DNN computation that boasts semi-honest security. This method leverages an efficient implementation of backpropagated gradients to replace quantization and normalization in secure computation without compromising accuracy. 

\textbf{Secret sharing:} Xu et al.~\cite{xu2019verifynet} adopted a variant of secret sharing technology.
The approach both safeguards the privacy of local gradients and addresses the issue of participants dropping out during training. Dong et al.~\cite{duan2020privacy} also incorporated secret sharing into the local shared gradients of participants in a method that allows the servers to both receive secret shares and compute the secret shares of aggregated gradients, striking a balance between efficiency and security. Applications of secret sharing extends to a federated transfer learning frameworks~\cite{sharma2019secure}. Here, the goal is to preserve data privacy against adversaries while also improving efficiency.

\subsubsection{Secure aggregation}
Secure aggregation protocols, which reside on the server of a federated learning system, combine the many local models sent by clients into a global model~\cite{bonawitz2017practical}. Within this protocol, each user conceals their local updates via a random mask before transmitting the masked updates to the server. The introduced randomness offsets both the private key and a paired random key, enabling the server to aggregate all the client models. Once the protocol has run its course, the server only has knowledge of the aggregated model, and not any knowledge of the individual models, as they remain obscured by unknown random keys.

So et al.~\cite{so2020byzantine} were the first to introduced the idea of a secure aggregation framework, and, notably, their implementation not only ensures privacy but also guarantees convergence. However, this does highlight that there is a fundamental trade-off between the scale of the network, user loss, and privacy protection. Elkordy and Avestimehr~\cite{elkordy2022heterosag} safeguard each user’s model updates by nullifying the mutual information between the masked model and the unshielded model. In this way, they secure the local model update in the strong information-theoretic sense. Heterogeneous quantization has also been deployed to strike a better balance between training accuracy and communication efficiency.

\subsubsection{Knowledge distillation}
Federated learning generally involves a multitude of decentralized computing nodes collaborating to train a centralized machine learning model, all while keeping local data samples local. This decentralized approach naturally results in uneven or disparate data distributions among the local models, compounded by multiple rounds of communication between nodes. This not only consumes substantial bandwidth but also amplifies the risk of leaks and introduces privacy challenges. To address these concerns, some privacy-preserving techniques employ distillation to further compress the size of the resultant global model while affording additional privacy assurances.

Sui et al.~\cite{sui2020feded} introduced a privacy-preserving medical extraction model tailored to the field of medicine that employs a knowledge distillation strategy. This strategy leverages the uploaded integrated local model for predictions, obviating the need to transmit the parameters of the entire local model to the central server. Another distillation approach is one-shot distillation, developed in response to the growing difficulty of servers and clients sharing large datasets. Gong et al.~\cite{gong2021ensemble} also proposed a federated learning framework tailored for one-shot distillation. This method safeguards the privacy of local data by exclusively utilizing model outputs devoid of any labels linked to common data during distillation-all without the need to exchange local model gradients. Furthermore, building upon prior research efforts, Gong et al.~\cite{gong2022preserving} also explored an offline strategy to bolster privacy that effectively disconnects the local training data from the local server.

\subsection{Privacy-preserving applications}
Privacy preservation in federated learning finds diverse applications in fields such as IoT, mobile edge computing, and blockchain. In this section, we offer an overview of the privacy-preserving methods for federated learning in each of these domains.

\subsubsection{Privacy-preserving in IoT}

Typically, IoT systems generate copious amounts of data, and traditional machine learning methods expose these data to privacy risks. However, endeavors to safeguard the privacy of this generated data within federated learning systems also need to mitigate the cost of data transmission. Federated learning finds utility across a spectrum of IoT domains, including smart homes~\cite{zhao2020privacy}, industrial settings~\cite{fu2020vfl}, power management~\cite{cao2020ifed}, and marine applications~\cite{qin2021privacy}, and these applications generally involve critically sensitive information that needs to be protected. Hence, privacy-preserving technology not only plays a vital role in safeguarding these systems but also in enhancing the efficiency and accuracy of these applications. Notably, in industrial IoT applications, federated learning can incorporate blind technology to shield participants’ gradient information
~\cite{fu2020vfl}. This ensures that the encrypted gradients of fellow participants remain secure and cannot be reversed.

\textbf{Power IoT:} Power IoT is a rapidly growing segment within IoT systems that is facing escalating privacy risks. Unauthorized access could potentially lead to the inference of household behavior patterns by reverse-engineering device statuses from energy consumption data~\cite{cao2020ifed}. In this context, local differential privacy protection is deemed more dependable than centralized protection. Federated learning, when applied to IoT scenarios, tackles the challenges posed when data is communicated to the cloud and provides robust support for privacy preservation while curbing communication costs.

\textbf{Mobile edge computing:} Traditional machine learning methods that rely on cloud-based infrastructure often centralize those data in cloud servers or data centers, raising critical privacy concerns. In response, edge computing has emerged as a promising solution, enabling the deployment of intelligent processes closer to the network's edge while maintaining control over personal data. Federated learning offers an effective means to collaboratively train machine learning models without the need to transmit raw data to the cloud. However, as more participants willingly engage in cooperative model training, the potential to jointly infer a model rises~\cite{lim2020federated}. Yet, when integrated with differential privacy, federated learning systems based on mobile edge computing can effectively thwart adversaries attempting to exploit network interactions to gain access to private edge data~\cite{gottipati2021fedran, zhang2020fedmec}. Still several challenges remain – significantly, resource allocation, computation costs, and handling unbalanced data in mobile edge computing.

\textbf{Blockchain:} Blockchain, characterized by its decentralized structure, encompasses both licensed and unlicensed blockchains. It is a unique data structure that aggregates chronological data blocks, with nodes to collectively manage and share data. Recently, there has been a surge of interest in harnessing blockchain technology for deep learning~\cite{weng2019deepchain}. Moreover, several novel applications have integrated federated learning into blockchain frameworks to protect privacy and as a way of offering incentives~\cite{shayan2020biscotti, awan2019poster, feng2021bafl}. For instance, Shayan et al.~\cite{shayan2020biscotti} introduced a decentralized framework called Biscotti that ingeniously melds blockchain and cryptography to safeguard client updates while maintaining global model performance, scalability, and fault tolerance. Concurrently, Awan et al.~\cite{awan2019poster} devised a blockchain-based privacy-preserving framework for federated learning settings, premised on the assumption that all the clients are semi-honest. Their method leverages blockchain's immutable and decentralized trust properties to serve as a reliable source for model updates. Overall, blockchain's intrinsic attributes,  i.e., that it is tamper-proof, cannot be falsified, and carries no repudiation, are instrumental in documenting the flow of information in federated learning scenarios.

\textbf{Smart healthcare:} Smart healthcare systems have traditionally relied on cloud sharing methods, which can make health information vulnerable to privacy attacks. However, health-related information is usually highly sensitive~\cite{act1996health}. In this context, federated learning has emerged as a promising solution for safeguarding privacy. For example, Choudhury et al.~\cite{choudhury2019differential} proposed a collaborative mobile phone and server training model designed for health monitoring that recognizes human activity. The application also includes an awareness of privacy protection measures. Transfer learning techniques are used to personalize the training model, taking the substantial variances observed among different individuals into consideration~\cite{chen2020fedhealth}. Li et al.~\cite{li2019privacy} introduced a federated learning model designed for federated brain imaging that segments brain tumors using a DNN. In this model, each client shares weight updates with the server for aggregation via a model averaging technique.

\section{Security}\label{sec:security}

\subsection{Security attacks}
In this section, we undertake a comprehensive classification of the security issues intrinsic to federated learning. These issues are discussed from four distinct perspectives: the models, the gradients, the training data, and the adversary’s perspective. Each section describes the various methods used to attack these components. This is followed by a discussion of the available methods to defend against these attacks.

\begin{figure}[h]
 \centering
 \includegraphics[scale=0.6]{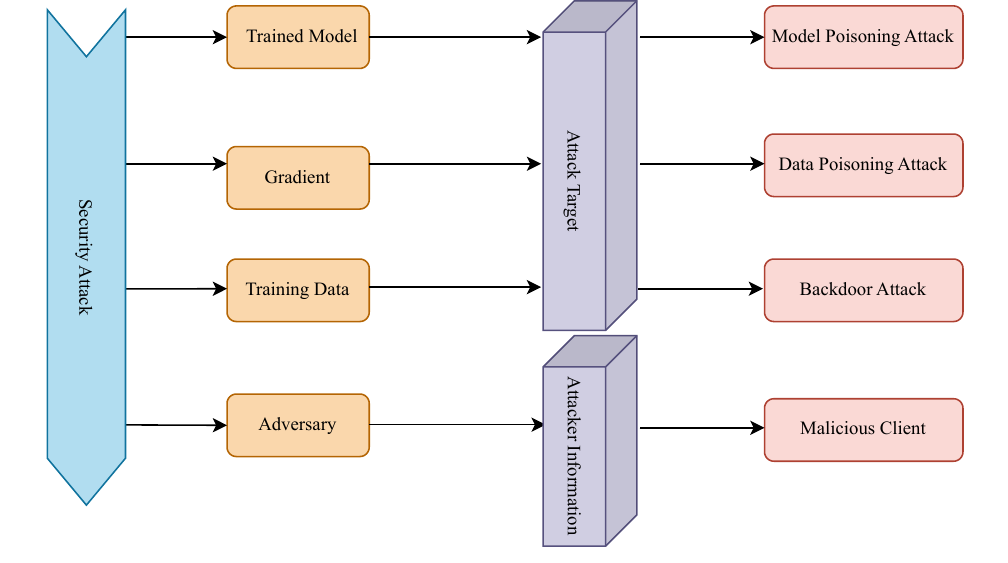}
 \caption{Security in federated learning different perspectives of attacks}
 \label{fig:security_issues}
\end{figure}

\begin{table}[]
\caption{The link between security issues and fairness}
\begin{tabular}{llll}
\hline
Security issues &         & Fairness & Remark                                             \\ \hline
Training Model        & Attack  & Yes   & Aggravates unfairness for clients  \\ 
               & Defense &  Yes        &  Increases unfairness in the trained model                                                  \\
Training Data     & Attack  &    Yes      & Increases  data heterogeneity, aggravates unfairness
                                                   \\
               & Defense &   Yes       &  Increases client fairness                                                  \\
Gradient       & Attack  &  Yes         &  Reduces fairness between clients                                                  \\
               & Defense &   Yes       &   Damages fairness for clients        \\ \hline                                         
\end{tabular}
\end{table}

\subsubsection{Attacks on trained models.} 
There are two main types of attacks on trained models, these being poisoning attacks and backdoor attacks. In either case, when a trained model is manipulated, either directly or indirectly, it can jeopardize the security of the entire training process. 

In a poisoning attack, the adversary can either attack the model or the data, with the aim of manipulating the target to sabotage its integrity. Further, in the case of a model poisoning attack, the attack can either be targeted or untargeted. Targeted attacks seek to minimize accuracy on specific test inputs, while untargeted attacks indiscriminately reduce accuracy on any test input within the global model.

Backdoor attacks, on the other hand, involve inserting hidden triggers into the training model that force the model to output some fixed response when tested. Our primary focus here is on instances where the trained model is directly affected, as is the case with both backdoor and poisoning attacks.

Shejwalkar and Houmansadr~\cite{shejwalkar2021manipulating} introduced a comprehensive framework for launching model poisoning attacks in a federated learning setting. The framework yields a multitude of poisoning attacks that surpass the capabilities of today's state-of-the-art model poison attacks. Adversaries can compute a malicious model update by perturbing a benign model aggregation with a malicious vector. Here, the goal is to avoid being detected by an aggregation algorithm. Fang et al.~\cite{fang2020local} proposed an attack that poisons local models, which involves manipulating the local parameters on a compromised device during the learning process. The objective is to construct local models on the compromised device in a way that maximally biases the global model in a specific direction, altering it before the attack occurs. Bhagoji et al.~\cite{bhagoji2019analyzing} devised a targeted model poisoning attack that introduces stealth. This covert attack blends malicious updates into an adversarial target.  Bagdasaryan ~\cite{bagdasaryan2020backdoor} designed a method of poisoning models that is based on replacing the model. The attacker use backdoor input data to train their model, with each training batch containing both correctly labeled data and backdoor data to facilitate the model distinguish between the two.

In all the above studies, both poisoning attacks and backdoor attacks have the potential to disrupt the availability of the federated learning system simply by compromising the trained model. Consequently, safeguarding the model is of paramount importance when attempting to build a robust security system.

\subsubsection{Attacks on data.}
When the integrity of a client’s training data is compromised, it has a cascading effect on the test error rate, ultimately putting a dent the global model’s accuracy. Adversaries typically undermine the veracity of training samples through tactics like data poisoning or backdoor attacks. Data poisoning involves directly and maliciously manipulating the training data, which has the potential to disrupt the entire system. Backdoor attacks on data are a form of data poisoning, with various schemes being devised for both image classification and text prediction~\cite{bagdasaryan2020backdoor}.

One prevalent method of data poisoning is label flipping. For example, Shejwalkar et al.~\cite{shejwalkar2022back} introduced a classic label flipping data poisoning attack based on label flipping and formulated the first comprehensive strategy to systematically address the data poisoning threat in federated learning. Tolpegin et al.~\cite{tolpegin2020data} similarly explored a label flipping attack as a way of executing a targeted data poisoning in federated learning. The effectiveness of such attacks hinges on the proportion of malicious participants in later rounds. Here, achieving maximum poisoning impact requires a high level of availability.

Backdoor attacks usually lead to the global model misclassifying specific test samples. Wang et al.~\cite{wang2020attack} outline how edge-case backdoors can lead to critical failures with substantial implications for fairness. Nguyen et al.~\cite{nguyen2020poisoning} devised a novel data poisoning attack that enables attackers to embed backdoors into an aggregation detection model, As a result, malicious traffic is misclassified as benign. Notably, attackers can leverage compromised IoT devices to implant backdoors without directly attacking the client. However, although these backdoors lack durability, they can persist in the model even after the attacker ceases to upload toxic updates. Zhang et al.~\cite{zhang2022neurotoxin} introduced Neurotoxin, a mechanism designed to embed more enduring backdoors into the training data within the federated learning systems. Notably, Neurotoxin can intelligently select the update direction of the update to circumvent clashes with benign clients.

From training data perspective, when some data is subject to malicious tampering or becomes tainted due to a poisoning or backdoor attack, it has the power to manipulate the global model. This not only endangers the security of the FL training process but also impinges upon the fairness for clients.

\subsubsection{Attacks on gradients.} 
Gradients serve as a critical intermediary element between clients and servers within federated learning frameworks. From a gradient perspective, both direct and indirect attacks have the potential to impact all the participants of a federated learning system. Additionally, it is important to note that attacks targeting both a trained model and the data can indirectly reverberate influencing the gradient. For instance, when a model itself is compromised or poisoned, the resulting gradient will carry this tainted information forward to the client or server, depending on the direction of the update. Similarly, data poisoning, which contaminates the training data, can also have an indirect effect on the gradient as this poisoned data is used for training, and will give rise to problematic gradients.

Both poisoning and backdoor implantation mechanisms fundamentally disrupt the updates generated by clients, with the overarching objective of altering the global model and compromising the security of the entire system. In the wake of model poisoning~\cite{shejwalkar2022back}, malicious gradients will propagate an increased amount of inverted label data with a deleterious effect on the overall accuracy of the global model. Further, a malicious subset  of mislabeled data will result in gradients that pollute the global model~\cite{tolpegin2020data}. Along these lines, Zhang et al.~\cite{zhang2022neurotoxin} introduced a more enduring form of backdoor attack relies on the attacker's ability to access the gradient in the preceding round, This means the adversary can approximate a benign gradient in the subsequent round by computing the first $k$\% coordinates of the benign gradient and strategically adjusting it according to a predefined constraint set.

In light of the above research, it is evident that adversaries do have the ability to compromise gradients through poisoning or backdoor attacks. The repercussions of such actions then cascade, impacting the global model and ultimately resulting in compromised security. Consequently, devising robust methodologies to mitigate the risks posed by backdoor and poisoning attacks has emerged as a formidable challenge in the domain of federated learning security.

\subsubsection{The adversary’s perspective}
Within the context of federated learning, adversaries can manifest as malicious clients that either infiltrate the system or work to compromise existing clients. These malevolent actors are driven by the overarching objective of sabotaging the performance of the global model. Hence, they pose a significant threat to the integrity of the entire federated learning framework.

Inherently, federated learning is a collaborative learning scheme, where clients collectively acquire shared global models. However, malicious clients can wield their influence to subvert the model selection process, causing the global model to predict spurious labels~\cite{cao2021provably}. As an example, consider an image classification scenario where a malicious client clandestinely relabels the training images of cars as birds.The client’s model update is then sent to the server. As a result, the learned global model is duped into classifying cars as birds~\cite{bagdasaryan2020backdoor}.

Malicious clients also have the ability to manipulate the direction and magnitude of local model updates~\cite{cao2020fltrust}. In this disconcerting scenario, each client represents a potential locus of both malice and susceptibility to poisoning attacks, as viewed from the server's vantage point. Notably, various studies~\cite{bhagoji2019analyzing,fang2020local,xie2019dba} have underscored that tampering with the model updates transmitted to the server can profoundly undermine the global model's Byzantine robustness during the learning process.

In a distinct vein, Sun et al.’s research~\cite{sun2021fl} introduces the concept of malicious clients sharing a dataset replete with corrupted data labels. Consequently, when this dataset is employed as input, the global model consistently produces these same corrupted labels as outputs. Furthermore, the malicious client propagates training data infused with these spurious labels~\cite{kim2022fedcc}, giving rise to pernicious learning models. This, in turn, precipitates a substantial deterioration in global learning outcomes.

Such declines in a global model's  performance are largely due to an adversary manipulating  the local client’s model or their local data. Effectively defending against these malevolent clients to uphold the security of federated learning constitutes a formidable challenge.

\subsection{Security defenses}
Table~\ref{tab:security} shows the correlation between security attacks and the corresponding defense methodologies, noting that the best choice of defense method is almost always contingent upon a number of factors, such as the nature of the attack and any known information about the attacker.

In a model poisoning attack, the adversary either tampers with the model or its parameters. Hence, the primary defense strategy revolves around identifying and removing these malevolent models. Conversely, when dealing with a data poisoning attack, the adversary’s goal is to manipulate or inject malicious content into the training data. Mitigating this attack involves eliminating the compromised or malicious party.

In scenarios involving backdoor attacks, the adversary clandestinely embeds a trigger in the model or training data. Defense mechanisms here encompass strategies like data perturbation or implementing a detection mechanism to identify and neutralize these surreptitious backdoors.

\begin{table}[h]
\Large
\centering
\caption{Link defense methods and security attacks}
\label{tab:security}
\renewcommand{\arraystretch}{2}
\resizebox{\linewidth}{!}{
\begin{tabular}{c|c|c|c|c|c}
\hline
References & MPA & DPA & BA & Attacker Information & Defense Method \\ \hline
\cite{shejwalkar2021manipulating} & \ding{52} &  &  & Malicious clients is craft malicious gradients & Detect outliers detection and removal-based spectral methods  \\
\cite{fang2020local} & \ding{52} &  &  & The adversary arbitrarily manipulates the local
models &Detect and remove the malicious local models  \\
\cite{zhao2022fedinv} &\ding{52}  &  &  &The adversary crafts a poisoned model that is similar to the benign model &Detect anomalous model updates and remove them \\
\cite{sun2021fl} & \ding{52} &  &  &Malicious devices perform the local training in different manners  &Perturb the parameters of the model  \\ \hline
\cite{jagielski2018manipulating} &  & \ding{52} &  &  The adversary injects poisoned data into the training set &The TRIM defense algorithm  \\ 
\cite{tolpegin2020data} &  & \ding{52} &  &Each malicious participant manipulates the training data& Detect anomalous updates and identify malicious participants \\ \hline
\cite{bagdasaryan2020backdoor} &  &  & \ding{52} &The adversary attempts to substitute the new global model with a malicious model  & Limit the updates by adding noise  \\
\cite{sun2019can} &  &  & \ding{52} &The adversary attempts to replace the whole model with a model that has a backdoor  &Apply norm thresholding to the updates or differential privacy  \\
\cite{rieger2022deepsight} &  &  & \ding{52} &The adversary injects a backdoor into the aggregated model &Detect the deep model with a filtering framework \\
\cite{nguyen2022flame} &  &  & \ding{52} &The adversary manipulates the global model  &Adding noise to the model updates. Identify and eliminate poisoned model updates. Clip the model weights before aggregation \\ \hline

\end{tabular}
}
\end{table}

\subsubsection{Anomaly detection}
The goal with anomaly detection is to identify any aberrant model updates or outliers training data, as these are the most likely candidates to be contaminated in a poisoning attack. Once detected, the aberrant update or outlying data are removed.

Muñoz et al. ~\cite{munoz2019byzantine} devised a method of detecting anomalies, attacks, and spurious updates within collaborative models based on adaptive federated averaging. The method leverages a hidden Markov model that assesses the quality of the client update and then discards any flawed or malevolent updates in each iteration of training. By contrast, Rieger et al.~\cite{rieger2022deepsight} introduced DeepSight, a filtering framework designed to counter backdoor attacks that filters out models with malicious intent. DeepSight combines a classifier with a clustering-based similarity estimation and probabilistic voting as its core mechanism. This approach mitigates the risk of benign client models with deviations in data distribution being erroneously filtered, which helps to maintain the integrity of the model aggregation process.

Shejwalkar and Houmansadr~\cite{shejwalkar2021manipulating} introduced a defense mechanism against model poisoning known as Dive-and-Conquer (DnC). DnC first computes the principal component of the input update set, then calculates the scalar product of each submitted model update with respect to this principal component, ultimately eliminating the constant component from the maximum scalar product within the submitted model update. Thus, DnC performs a spectral analysis of the input updates through dimension reduction to ensure that malicious updates are detected.

Detecting anomalous client updates can also help to pinpoint their adverse impacts.  For example, Jagielski et al.~\cite{jagielski2018manipulating} proposed the TRIM defense algorithm, which is designed to combat data poisoning attacks. TRIM endeavors to identify the model parameters and a subset of the training set (of a predefined size) that will work to minimize the loss function. To this end, it iteratively estimates the regression 
parameters while employing a constructed loss function to eliminate data points with substantial residuals. Fang et al.~\cite{fang2020local} devised two Byzantine-robust federated learning defense strategies against poisoning attacks. The first strategy is to evaluate the effect that the local models will have on the error rate of the validation dataset and then discard any local models that have a significant adverse impact. The second shifts the local model based on its impact on loss rather than the validation set error rate. These solutions appear to work well against specific adversary models as they follow assumptions about the adversary’s attack strategy and the underlying distributions of the dataset, whether benign or adversarial. However, if these assumptions do not hold, these defense strategies may prove ineffective.

In addition to purging local model updates from clients, Cao et al.~\cite{cao2020fltrust} introduced the FLTrust framework, in which the service provider assigns trust scores to each local model update during each iteration. If the direction of local model updates significantly deviates from the update server model, the trust scores for these updates increase. Subsequently, the size of the local model updates is normalized to position them within the same hypersphere as the server model updates in the vector space. This limits the influence of malicious local model updates.

Zhao et al. ~\cite{zhao2022fedinv} proposed an alternative approach
to defending against poisoning attacks. Their solution is a new Byzantine-robust federated learning framework that inverts the local model updates. Before sending the global model update in each round, the server performs a model inversion on each client’s local update and synthesizes a corresponding virtual dataset. The Wasserstein distance~\cite{deshpande2019max} between the generated virtual cube and the others is then computed,  and any local update with a very large distance is not aggregated with the other local models.

\subsubsection{Differential privacy}
In the realm of safeguarding against model poisoning attacks, perturbations serve as a crucial defense mechanism. These perturbations can be introduced during local training or applied to the aggregated global model to shield clients from such attacks. The core objective of perturbation is to mitigate the impact of these malevolent attacks by introducing controlled noise, often through constraints on the updated $l_2$-norm \cite{bagdasaryan2020backdoor}.

In pursuit of a training model with differential privacy to thwart model poisoning attacks~\cite{sun2019can}, Sun et al. ~\cite{sun2021fl} proposed an innovative methodology. In this approach, the server identifies updates exceeding a predefined threshold of potential toxicity $M$, the assumption being that adversaries will be aware of the threshold $M$. This strategy effectively aligns the boundary defense norm with local clipping during updates. Following this step, differential privacy measures are applied to the aggregated global model, strengthening resilience against model poisoning attacks. Sun et al.~\cite{sun2021fl} further introduced the concept of Federated Learning-White Blood Cell (FL-WBC), a client-centric approach to combatting model poisoning attacks originating from a tainted global model. The method incorporates a quantitative parameter estimation technique to identify a parameter space with enduring effects on the local training parametersn that subsequently introduces perturbations to mitigate these effects.

Nguyen et al.~\cite{nguyen2022flame} devised the Flexible Learning Against Model Poisoning Attacks framework (FLAME), specifically designed to counter backdoor attacks. FLAME injects controlled noise into a model to neutralize any backdoors while employing model clustering and weight clipping techniques to minimize the noise necessary, all the while preserving benign performance. FLAME excels in negating the impact of backdoors. It accomplishes this by identifying and eliminating potentially toxic model updates through automated model clustering and implementing model weight pruning before aggregation to curb the influence of malicious model updates on the aggregation results. However, it is important to note that this defense mechanism has limitations, particularly in cases of poisoned models with high-impact attacks. When training samples exhibiting backdoor behavior are introduced into the original benign training data, the poisoned model will often yield higher accuracy on the backdoor task, rendering this defense less effective.

\subsection{Security applications}
\subsubsection{Anomaly detection in IoT}
Federated learning encompasses a wide array of participants, including IoT devices, where only the model parameters are exchanged between the cloud server and the clients. However, this FL framework is not immune to poisoning attacks initiated by internal programs \cite{zhang2020poisongan}. IoT networks, in particular, face the daunting challenge of accommodating a substantial number of potentially malicious clients, some of which may be harboring malicious users. Given that the user's local data and training process remain concealed from the server, validating the authenticity of a user's update can become a challenging endeavor. Additionally, IoT devices place a premium on energy efficiency, which renders the deployment of computationally intensive security firewalls impractical and, consequently, leaves them susceptible to various forms of attack.

To mitigate these security concerns, Mothukuri et al.~\cite{mothukuri2021federated} introduced an FL anomaly detection method designed to detect attacks and proactively identify intrusions within IoT networks. This method leverages data from distributed devices to facilitate on-device training that teaches machine learning models to detect anomalies within IoT networks – all without the need to transfer network data to centralized servers. To tackle the issue of tag noise, Chatterjee and Hanawal~\cite{chatterjee2022federated} proposed a tag noise intrusion detection system tailored to the security needs of IoT networks, capable of adapting to federated settings. Another noteworthy approach is the deployment of a decentralized asynchronous federated learning framework underpinned by blockchain technology within the IoT domain~\cite{cui2021security}. This framework achieves global aggregation in federated learning through blockchain consensus mechanisms rather than relying on a central server, thus providing a secure and accurate means of detecting anomalies within IoT networks.

Both attack detection and security protection are integral components of fortifying federated learning within IoT. However, it is important to recognize that the detection mechanisms may themselves be vulnerable to various attacks, including backdoor attacks that misclassify malicious traffic as benign. Addressing this challenge is paramount for enhancing the security of federated learning in IoT systems.

\subsubsection{Visual detection protection}
In the realm of traditional security, data acquisition often relies on camera systems, with monitoring rooms serving as the central hubs for observation. Nevertheless, this conventional approach does not consistently guarantee security~\cite{zheng2022applications}. It is susceptible to shortcomings – for example, where anomalies in the behavior of personnel may not be noticed in a timely manner, resulting in missed warnings and erroneous assessments.

In response to these challenges, Liu et al.~\cite{liu2020fedvision} proposed an innovative solution aimed at bolstering security monitoring in smart cities. Rooted in federated learning and visual detection, this novel approach effectively harnesses data from multiple communities to construct a resilient security model. It interconnects different communities and exchanges information between them. As such, it gives rise to a myriad of potential applications, including but not limited to fire monitoring.

On the whole, this approach emerges as a promising remedy for the limitations inherent to traditional security paradigms that rely on centralized monitoring. That said, further research is needed to evaluate its effectiveness and potential limitations in a diverse range of contexts.

\section{Challenges and Open Research Directions in Federated Learning}\label{sec:challenge}
This section is dedicated to providing insights into some of the fundamental challenges encountered in federated learning. While federated learning holds immense promise for safeguarding the privacy of clients’ local data, there are still a multitude of privacy, security, and fairness concerns to grapple with. These include issues such as privacy breaches, security vulnerabilities, and disparities in fairness. Resolving these challenges and exploring future trajectories in the field is therefore highly worthy of a comprehensive examination.

\subsection{The Impact of Privacy and Security on Fairness}
Bias in machine learning models is a persistent concern, capable of significantly shaping decision-making processes and potentially culminating in unfair outcomes. In the realm of federated learning, the challenge of bias takes on heightened complexity due to the inherent variability of client data. As such, it is imperative to consider fairness alongside privacy and security in the context of federated learning~\cite{kairouz2021advances}. Part of this requires acknowledging that safeguarding privacy and ensuring security can mean unintended bias might surface.

Take, for example, the application of differential privacy measures, which, while fortifying privacy, may inadvertently amplify unfairness. Similarly, in the process of filtering out malicious adversaries as part of security measures, there is risk of erroneously identifying benign clients as potential attackers. Such misclassifications often arise because the nature and distribution of that client’s data departs from the majority. This is a form of discrimination. Consequently, it is paramount to explore the repercussions of privacy and security on fairness and implement measures to effectively curtail bias.

In future research on federated learning, two promising topics for investigation include how to cultivate trust and how to tailor personalized methodologies that might ameliorate the impact of privacy and security on fairness. In-depth explorations of the intricate interplays between privacy, security, and fairness, might give rise to targeted approaches along these lines that both mitigate bias and advance the cause of equity within federated learning.

\subsection{The balances between privacy, security and fairness}
Examining the intricate equilibrium between privacy, security, and fairness within the realm of federated learning stands as a subject of paramount significance. Historically, research endeavors have predominantly gravitated toward investigating these three pillars in isolation, with some dedicating their efforts solely to the preservation of privacy and security within federated learning, while others have delved into rectifying fairness concerns among clients or cohorts. However, recent scholarship has underscored the intrinsic interconnections between privacy, security, and fairness ~\cite{cummings2019compatibility,gu2022privacy, ozdayi2021impact}. Recognizing and harnessing these intricate connections could catalyze the development of methodologies that concurrently elevate fairness, safeguard privacy, and fortify security within federated learning.

As an example, empirical evidence suggests that ensuring fairness in federated learning might be achievable if one could identify and exclude malicious contributors and adversaries. In turn, this would help to bolster privacy and security~\cite{xu2020reputation}. Further, exploring novel approaches to ensuring fairness, privacy, and security in concert, such as identifying and commending clients based on the fidelity of their submitted gradients, offers a multifaceted avenue for addressing concerns related to cyber safety. Consequently, researchers in the field must begin to pursue a balanced and holistic approach to safeguarding data – one that takes into account the inherent interconnectedness of privacy, security, and fairness. This is the grail that holds the most significant promise in surmounting the complexities posed by these challenges.

By embracing the intricate tapestry of relationships and diligently striving for equilibrium among privacy, security, and fairness, researchers are poised to make substantial strides in conquering these formidable obstacles. The judicious consideration and harmonization of these dimensions herald a future characterized by comprehensive solutions that not only champion fairness but also steadfastly uphold the imperatives of privacy and security.

\section{Conclusion}\label{sec:conclusion}
Prior research studies ~\cite{abdulrahman2020survey,blanco2021achieving,gosselin2022privacy,mothukuri2021survey,truong2021privacy,zhang2022challenges,shen2022distributed} have exhaustively investigated the discrete realms of privacy, security, and fairness within the paradigm of federated learning, with numerous researchers proposing various mechanisms of defense to safeguard these modalities. However, these undertakings have predominantly fixated on redressing privacy, security, or fairness in isolation. To navigate these multifaceted challenges adeptly, it is paramount to explore the foundational interrelations that underpin these three constructs.

In this study, we offer pioneering insights into the intricate tapestry of connections binding privacy, security, and fairness, ushering forth a fresh perspective predicated on achieving equilibrium between the three. We find that privacy and security concerns can intertwine, most notably through the conduit of gradient sharing, and that a trade-off dynamic exists between most combinations of these dimensions. These revelations substantially broaden our comprehension of the forthcoming challenges awaiting federated learning within the domains of privacy, security, and fairness, proffering innovative avenues for resolving these intricate quandaries. Significantly, our inquiry introduces the novel concept of fairness as a unifying bridge that links the realms of privacy and security, offering a transformative outlook on this dynamic.

Consequently, our study underscores the imperative of perceiving privacy, security, and fairness as interconnected facets rather than solitary predicaments when grappling with the intricacies of privacy or security within federated learning. This is a perspective that also underscores the need to holistically contemplate privacy, security, and fairness in any federated learning system. To the best of our knowledge, this is the first study to deeply explore the nexus binding privacy, security, and fairness. For this reason, we recommend that subsequent research endeavors be tailored toward applying these discoveries to real-world federated learning scenarios, particularly those that involve numerous clients. In addition to addressing the security and privacy requisites of these multiple parties, treating clients equitably is an indispensable consideration for the path forward.

\begin{acks}
This research is supported by the NSFC-FDCT as part of the Joint Scientific Research Project Fund (Grant No. 0051/2022/AFJ), China \& Macau.
\end{acks}

\bibliographystyle{ACM-Reference-Format}
\bibliography{main}

\end{document}